\documentclass[11pt]{article}
\usepackage[marginratio=1:1,height=600pt,width=460pt,tmargin=100pt]{geometry}
\usepackage{authblk}

\usepackage[utf8]{inputenc} % allow utf-8 input
\usepackage[T1]{fontenc}    % use 8-bit T1 fonts
\usepackage{hyperref}       % hyperlinks
\usepackage{url}            % simple URL typesetting
\usepackage{booktabs}       % professional-quality tables
\usepackage{amsmath,amsthm,amssymb, amsfonts}       % blackboard math symbols
\usepackage{nicefrac}       % compact symbols for 1/2, etc.
\usepackage{microtype}      % microtypography
\usepackage{xcolor}         % colors

\usepackage[style=ieee]{biblatex}
\usepackage[super]{nth}
\usepackage{bm}
\usepackage{csquotes}
\usepackage{subcaption}
\usepackage{algorithm}
\usepackage{algpseudocode}

\usepackage{titlesec}
\titleformat*{\paragraph}{\itshape}

\title{Nonlinear Time-Series Embedding by Monotone Variational Inequality}

\author[1]{Jonathan Y. ~Zhou}
\author[1]{Yao Xie}
\affil[1]{{\small H. Milton Stewart School of Industrial and Systems Engineering, Georgia Institute of Technology.}}

\usepackage[colorinlistoftodos,prependcaption,textsize=tiny]{todonotes}

\usepackage[acronym,nohypertypes={acronym,notation}]{glossaries}

\usepackage[english]{babel}

\DeclareMathOperator{\diag}{diag} % diagonal
\DeclareMathOperator{\vc}{vec} % vectorize
\DeclareMathOperator{\Prox}{Prox} %proximal
\DeclareMathOperator{\spanl}{sp} %linear span

\DeclareMathOperator*{\argmin}{arg\,min}

\newcommand{\mtx}[1]{\mathbf{#1}} % Matrix
\newcommand{\op}[1]{\mathcal{#1}} % Linear Operator

\newacronym{ista}{ISTA}{Iterative Shrinkage-Thresholding Algorithms}
\newacronym{seqcluster}{SEQEm}{\textbf{Our Method of Sequence Embedding}}
\newacronym[firstplural={Variational Inequalities}, longplural={Variational Inequalities}]{vi}{VI}{Variational Inequality}
\newacronym{glm}{GLM}{Generalized Linear Model}
\newacronym{mimic-ii}{MIMIC-II}{Multiparameter Intelligent Monitoring in Intensive Care II}
\newacronym{nmi}{NMI}{Normalized Mutual Information}
\newacronym{fista}{FISTA}{Fast Iterative Shrinkage and Thresholding Algorithm}
\newacronym{saa}{SAA}{Sample Average Approximation}
\newacronym{var}{VAR}{Vector Autoregression}
\newacronym{iid}{i.i.d.}{Independent and Identically Distributed}
\newacronym{stpp}{STPP}{Spatio-Temporal Point Processes}
\newacronym{nspp}{NSPP}{Non-Stationary Point Processes}
\newacronym{mle}{MLE}{Maximum Likelihood Estimation}
\newacronym{svd}{SVD}{Singular Value Decomposition}
\newacronym{sgd}{SGD}{Stochastic Gradient Descent}
\newacronym{gd}{GD}{Gradient Descent}
\newacronym{pca}{PCA}{Principal Component Analysis}
\newacronym{dbscan}{DBSCAN}{Density-based spatial clustering of applications with noise}
\newacronym{svm}{SVM}{Support Vector Machine}
\newacronym{dtw}{DTW}{Dynamic Time Warping}
\newacronym{llm}{LLM}{Large Language Model}
\newacronym{tsne}{T-SNE}{t-Distributed Stochastic Neighbor Embedding}
\newacronym{ari}{ARI}{Adjusted Rand Index}
\newacronym{cp}{CP}{Clustering Purity}
\newacronym{edat}{EDAT}{Empirical Distribution of Inter-Arrival Times}
\newacronym{icu}{ICU}{Intensive Care Unit}
\newacronym{em}{EM}{Expectation Maximization}
\newacronym{adam}{ADAM}{Adaptive Moment Estimation}
\newacronym{socp}{SOCP}{Second-Order Cone Program}
\newacronym{ekg}{ECG}{electrocardiogram}
\newacronym{sa}{SA}{Stochastic Approximation}
\newacronym{ls}{LS}{Least Squares}
\newacronym{md}{MD}{Mirror Descent}
\newacronym{lda}{lda}{Latent Dirichlet Allocation}
\newacronym{hmm}{HMM}{Hidden Markov Model}
\newacronym{svt}{SVT}{Singular Value Thresholding}
\newacronym{erm}{ERM}{Empirical Risk Minimization}
\newacronym{arima}{ARIMA}{Autoregressive Integrated Moving Average}
\newacronym{iot}{IOT}{Internet of Things}
\newacronym{sdp}{SDP}{Semidefinite Program}
\newacronym{knn}{KNN}{K-nearest neighbors}
\newacronym{dgf}{DGF}{Distance Generating Function}

\begin{document}

\maketitle
\begin{abstract}
In the wild, we often encounter collections of sequential data such as electrocardiograms, motion capture, genomes, and natural language, and sequences may be multichannel or symbolic with nonlinear dynamics. We introduce a new method to learn low-dimensional representations of nonlinear time series without supervision and can have provable recovery guarantees. The learned representation can be used for downstream machine-learning tasks such as clustering and classification. The method is based on the assumption that the observed sequences arise from a common domain, but each sequence obeys its own autoregressive models that are related to each other through low-rank regularization. We cast the problem as a computationally efficient convex matrix parameter recovery problem using monotone \glspl{vi} and encode the common domain assumption via low-rank constraint across the learned representations, which can learn the geometry for the entire domain as well as faithful representations for the dynamics of each individual sequence using the domain information in totality. We show the competitive performance of our method on real-world time-series data with the baselines and demonstrate its effectiveness for symbolic text modeling and RNA sequence clustering.
\end{abstract}

\section{Introduction}

Collections of time-series data, where each sequence is represented by a series of points indexed over time, are ubiquitous and increasingly prevalent. Notable examples include physiological signals \cite{10.7551/mitpress/9609.001.0001, PerezAlday_2020}, power systems \cite{801851}, financial data \cite{GVK483463442, DBLP:journals/corr/abs-2101-04285}, computer networks \cite{493355}, and electronic health records \cite{Reyna2020-ph, medbert}. In addition to traditional time series data, other sequential data like genes and proteins  \cite{Argelaguet2020, protiens} as well as natural language have garnered significant recent attention, particularly with the advent of large language models \cite{brown-etal-1992-class, peters-etal-2018-deep, reimers-gurevych-2019-sentence,cer-etal-2018-universal}. 

Learning high-quality representations of sequences and time series \cite{NIPS2013_9aa42b31} is an essential building block \cite{10.1109/TPAMI.2013.50} and important for understanding the dynamics underlying observed sequences, enabling informed decisions and downstream machine learning tasks \cite{trirat2024universal}. A key paradigm underlying the unsupervised representation learning has been that of self-supervised learning \cite{e26030252}, where we first solve some auxiliary task (e.g., autoregression or reconstruction), which leads implicitly to a compressed representation of the input data \cite{pml2Book, kingma2022autoencoding}. The development of self-supervised methods for natural language has been, in turn, paralleled by embedding methods for other types of sequential data, and there is now a burgeoning literature on time series and sequence representation \cite{trirat2024universal, Lafabregue:2022aa, medssl}.

A lasting challenge in bringing representation learning to time series is how to learn the information common to the entire domain in conjunction with faithful {\it individual} representations of each sequence. In contrast to the natural language setting where there exists a shared ``universal'' embedding space for all languages \cite{yang-etal-2020-multilingual} backed up by a large amount of available training data, time series data are often highly domain-specific in the sense that each domain is distinct from another (e.g., \gls{ekg} vs power generation). Additionally, the individual processes which we receive observations for may, in fact, be substantially different among themselves (e.g., differences between patients and power plants). From the above, there is a difficulty in balancing learning the common dynamics of a set of observed sequences in addition to faithfully representing the dynamics of each sequence. This is especially the case when observations of all sequences individually are limited. To this end, we take inspiration from an area where the above challenges are common and well known --- low-rank matrix recovery \cite{Davenport_2016, 5452187, 5454406, 6942195, statinferenceconvex} --- and develop it towards the general sequential and time series representation learning setting, enabling us to bring provable recovery guarantees to the modeling of a broad class of sequences with autoregressive character.

To this end, we introduce an approach to the unsupervised learning of low-dimensional representations for nonlinear time series based on the assumption that each sequence behaves according to its own autoregressive model, but that the sequences are related to each other through low-rank regularization. We cast the problem as a computationally efficient convex matrix parameter recovery problem using monotone \glspl{vi}. By enforcing the low-rank assumption across the learned representations, we capture the geometry of the entire domain and accurately represent the dynamics of each individual sequence utilizing the entire domain information. We apply our method to real-world time-series data and demonstrate its effectiveness in symbolic text modeling and RNA sequence clustering.

\subsection{Related work}
\label{sec:LR}
We first review related work in time-series representation learning, clustering, and classification. A number of methods exist to address the above problems, the simplest of which is to extract features \cite{10.1145/1557019.1557122} or establish a distance metric between time-series \cite{Cormen2001introduction, Muller2007, bakeoff}. Another approach is to construct a model for each series --- our work falls within this purview of model-based time-series representation \cite{NIPS1996_6a61d423, arima}. Most recent time-series representation learning approaches adopt a contrastive learning framework to disambiguate between sequences (e.g., by taking sub-samples of the same sequence as positive samples and from differing sequences as negative ones) using deep networks \cite{pmlr-v162-yang22e, Yue_Wang_Duan_Yang_Huang_Tong_Xu_2022, 10233880, fraikin2024trep, wang2023contrast}. Recent work in this line has focused particularly on the choice of neural architecture, the data augmentation to ensure robustness, and the selection scheme for contrastive learning. Other auxiliary tasks for self-supervised representation learning of time series include reconstruction and context prediction \cite{NEURIPS2019, fortuin2020gp}, also popular for language modeling \cite{devlin-etal-2019-bert}.  

In our work, we adopt auto-regression as the auxiliary task. Unlike recent methods that use contrastive learning to learn an encoder for the latent space indirectly, we do not assume inherent similarities or differences across the sequences. Instead, we directly constrain the geometry of the representations themselves to be low-rank. We motivate our work from the perspective of low-rank matrix recovery \cite{Davenport_2016}, common in other areas of machine learning and which serves as the foundation for principal component analysis \cite{hotelling1933analysis}, classical methods in natural language processing (topic modeling) \cite{10.5555/944919.944937, 10.1145/2133806.2133826} and collaborative filtering \cite{5197422}. Problems from this area often admit convex formulations and come with provable recovery guarantees \cite{statinferenceconvex}. Most recently, a line of work has loosened the structural assumptions needed for signal (time-series) recovery in an autoregressive context while still maintaining problem convexity, using \glspl{vi} with monotone operators the main tool \cite{juditsky2023generalized, 9273032, ssso}. However, to our knowledge, we are the first to apply these ideas in conjunction with low-rank matrix recovery for modeling and representing collections of general time-series observations.

\section{Problem Setup}

We aim to representing observations into $N$ vector-valued time series of length $T$ each of the form $\{\mathbf{x}_{i,t}\}$, where $\mathbf{x} \in \mathbb{R}^C$, $t \in [T]$, and $i \in [N]$. The sequences are sampled from a common domain independently of each other across $i$, but have temporal dependence across $t$. We refer to the history of events for sequence $i$ up to time-point $t$ as $\mathcal H_{i,t} := \{\mathbf{x}_{i,s} \mid s < t\}$. We expect the behavior at event $\mathbf{x}_{i,t}$ to be a function of past observations. In particular, for time-point $\mathbf{x}_{i,t}$, we suppose the dependence on $\mathcal{H}_{i,t}$ is sufficiently captured an order $d$ nonlinear vector autoregressive model with $C$ channels on $\bm{\xi}_{i,t} = \vc(1,\{\mathbf{x}_{i,t-s}\}_{s=1}^d) \in \mathbb{R}^{Cd+1}$, the values from the $d$ preceding observations and a bias term so that
\begin{equation}\label{eqn:obsmodel}
    \mathbb{E}[\mathbf{x}_{i,t} \mid \mathcal{H}_t] = \eta(\mathbf{R}_{i}\bm{\xi}_{i,t}).
 \end{equation}
The matrices $\mathbf{R}_i \in \mathbb{R}^{C \times (Cd+1)}$ each serve as weights for the prediction of the focal observation $\mathbf{x}_{i,t}$ that we aim to learn. We allow $\mathbf{b}_i = \vc({\mathbf{R}_i}) \in \mathbb{R}^{C^2d + C}$ to be the parameters corresponding to the $i^\text{th}$ sequence arranged as a vector and sufficient to capture the dynamics of the $i^\text{th}$ time-series. The function $\eta:\mathbb{R}^C \to \mathbb{R}^C$ is a monotone \textit{link function}. The choice of link function $\eta$ naturally corresponds to the character of the recovered sequence dynamics, which we illustrate via the following examples:
\begin{description}
    \item[Vector auto-regression]$\eta(\mathbf{x})=\mathbf{x}; \mathbf{x} \in \mathbb{R}^n$, e.g.\ motion capture, \acrfull{ekg} signals.
    \item[Symbolic sequences] $\eta(\mathbf{x})=\exp({\mathbf{x}})/\sum_{i}\exp(x_i); \mathbf{x}\in[\Sigma]^n$, e.g.\ natural language, genes.
    \item[Count processes] $\eta(\mathbf{x})=\exp(\mathbf{x}); \mathbf{x} \in \mathbb{Z}_{\geq 0}^n$, e.g.\ traffic intensity, call center arrival rates.
    \item[Bernoulli Processes] $\eta(\mathbf{x})= \exp(\mathbf{x})/(1+\exp(\mathbf{x})); \mathbf{x} \in \mathbb{B}^n$, e.g.\ wildfire presence, neuron firing.
\end{description}

We do not restrict the mechanics of the link function $\eta$ beyond the monotone property. We also remark that each vector $\mathbf{b}_i$ corresponding to each sequence may itself be high dimensional. The key aspect of our method for low dimensional representation learning lies in the common domain assumption, which should limit how the sequences are similar (different). We leverage this information by a \textit{low rank} assumption on the space of parameters by which each sequence is described. In this way, we constrain the individual $\mathbf{b}_i$ to lie approximately on a \textit{low dimensional subspace} of the possible full parameter space of $\mathbf{b}_i$s. The representation of each sequence within this subspace may be taken as an embedding and used for downstream tasks such as clustering, classification, and anomaly detection.

In particular, we consider the autoregressive sequence model introduced in  \eqref{eqn:obsmodel}, allowing $\mathbf{b}_i$ to be those parameters unique to the $i$\textsuperscript{th} sequence. Allow the matrix $\mtx{B} =\begin{bmatrix} \mathbf{b}_1 & \ldots & \mathbf{b}_N \end{bmatrix} \in \mathbb{R}^{m \times N}$ denote the parameters across all the sequences. We aim to recover a good choice of the matrix $\mtx{B}$ without supervision. We should balance two goals: (1) we desire each $\mathbf{b}_i$ to be as faithful to the generating dynamics of their respective observed data as possible; (2) we hope to leverage the \textit{common domain assumption} about the sequences and use information from the other sequences to inform the prediction of the focal sequence. To express the corresponding low-rank constraint, consider the rank $r$ \gls{svd} of $\mtx{B}$
\begin{equation}
  \mtx{B} = \mathbf{U}\mathbf{\Sigma}\mathbf{V^*} = \sum_{k=1}^r \sigma_k \mathbf{u}_k \mathbf{v}_k^*  
\end{equation}
where $\mathbf{\Sigma} = \diag\{\sigma_k\}_{k=1}^r$ corresponds to the singular values, columns of $\mathbf{U} = [\mathbf{u}_k]_{k=1}^r \in \mathbb{R}^{m \times r}$ form an orthobasis in $\mathbb{R}^d$, and columns of $\mathbf{V}^* = [\mathbf{v}_k]_{k=1}^r \in \mathbb{R}^{r \times N}$ form an orthobasis in $\mathbb{R}^N$. The recovered columns $\mathbf{C} := \mathbf{\Sigma} \mathbf{V}^* = [\mathbf{c}_i]_{i=1}^N \in \mathbb{R}^{r\times N}$ give an $r$-dimensional representation for each of the $N$ sequences. Likewise, the geometry of the subspace $\spanl\{\mathbf{u}_k\}_{k=1}^r$ describes the \textit{common domain} from which the generating processes of the sequences arise.  We consider the low dimensional representation for the $i$\textsuperscript{th} sequence $\mtx{b}_i = \mtx{U}\mtx{c}_i$ as the embedding of the dynamics for the $i^\text{th}$ sequence.

Because rank-constrained optimization is, in general, an NP-hard problem \cite{doi:10.1137/S0097539792240406}, to enforce the low-rank requirement on $\mtx{B}$, we instead constrain our setup to a \textit{nuclear norm ball}. The nuclear norm is given by $\|\mathbf{X}\|_* = \sum_{j=1}^r \sigma_i(\mathbf{X})$ where $\sigma_i$ is the $i^\text{th}$ singular value of the matrix $\mathbf{X}$. The nuclear norm is also the tightest convex relaxation to matrix rank \cite{doi:10.1137/070697835} leading to tractable parameter recovery and allows us to leverage a long line of work from convex optimization and matrix recovery \cite{doi:10.1137/080738970, Davenport_2016, Nesterov_Nemirovski_2013}.

In terms of the modeling of sequences themselves, 
we treat each sequence as arising from an individual stochastic source from which we acquire observations and aim to learn a sequence-level autoregressive model. We cast this parameter recovery problem as a \textit{monotone variational inequality}, the most general form of convex problem for which there exist efficient solution methods \cite{statinferenceconvex,facchinei_finite_dimensional_2003_1, ssso, 9273032}. The objective is designed so that the dynamics should be faithful in the sense that, for a focal sequence, the learned dynamics can be used to generate sequences similar to the focal time-series observations while still maintaining problem convexity.

\section{Method}

In the following, we present our method, first for linear auto-regressive models and then for general non-linear auto-regressive models, which can be applied to categorical sequences. 

\subsection{Low rank time-series embedding for linear auto-regressive models} \label{sec:AR.linear}
First, suppose events $\mathbf{x}_{i,t} \in \mathbb{R}^n$ obey a linear autoregressive model. Recall that we allow $\bm{\xi}_{i,t} = \vc(1,\{\mathbf{x}_{i,t-s}\}_{s=1}^d) \in \mathbb{R}^{Cd+1}$, the values from the $d$ preceding observations and bias term so that
\begin{equation}\label{eqn:univar.reg}
    \mathbb{E}[\mathbf{x}_{i,t} | \mathcal{H}_{t}] = \mathbf{R}_{i}\bm{\xi}_{i,t}.
\end{equation}
The matrices $\mathbf{R}_i \in \mathbb{R}^{C \times (Cd+1)}$ serve as weights for the prediction of the focal observation $\mathbf{x}_{i,t}$. Recall also that we allow $\mathbf{b}_i = \vc({\mathbf{R}_i}) \in \mathbb{R}^{C^2d + C}$. To recover the parameter matrix $\mtx{B}=[\mathbf{b}_i]_{i=1}^N \in \mathbb{R}^{(C^2d+ C) \times N}$, a natural choice is to take least squares loss and write
\begin{align}
\label{eqn:lin.b.lstq}
\min_{\widehat{\mathbf{B}} \in \mathbb{R}^{d \times N}} \frac{1}{N}\sum_{i=1}^N \left( \frac{1}{T-d}\sum_{t=d+1}^T \| \mathbf{x}_{i,t} - \widehat{\mathbf{R}}_i\bm{\xi}_{i,t} \|_2^2\right) \qquad \text{ s.t. } \qquad \|{\widehat{\mathbf{B}}}\|_* \leq \lambda.
\end{align}

\paragraph{Low-rank recovery and the nuclear norm regularization.}
We now discuss Program \eqref{eqn:lin.b.lstq} in the context of low-rank matrix recovery  \cite{Davenport_2016}. We aim to recover matrix $\mtx{B}$, but instead of observing it directly, we receive indirect samples $\mathbf{y} \approx \op{A}(\mtx{B})$ through random linear \textit{measurement operator} $\op{A}: \mathbb{R}^{(C^2d+ C) \times N} \to \mathbb{R}^{CN}$ such that $\mathbb{E}[\mathbf{y}|\mathcal{A}] = \mathcal{A}(\mathbf{y})$. Namely, we realize samples of form $(\op{A}_t, \mathbf{y}_t)$ by selecting a time-point $t$ across the observation horizon. The samples $ \mathbf{y}_t := \vc([\mathbf{x}_{i,t}]_{i=1}^N) \in \mathbb{R}^{CN}$ represent the values from the present time-point across all $N$ sequences. The sampled operator $\mathcal{A}_t$ packages together the preceding length $d$ window of regressors across the $N$ sequences and adds also the corresponding random noise $\bm{\epsilon}$ so that
\begin{equation}\label{eqn:op.var}
    \mathcal{A}_t(\widehat{\mathbf{B}}) := \frac{1}{N}\vc([\mathbf{R}_i \bm{\xi}_{i,t} + \bm{\epsilon}]_{i=1}^N) : \mathbb{R}^{(C^2d+C) \times N}\to \mathbb{R}^{CN}.
\end{equation}
We take noisy indirect observations $\mtx{y}\approx \op{A}(\mtx{B})$ to be drawn from $(\mathcal{A}, \mtx{y}) \sim P$ and consider the expected least squares loss via stochastic program 
\begin{equation}
    \label{eqn:lowrankrecoverynuclear}
    \min_{\widehat{\mathbf{B}}} \ell(\widehat{\mathbf{B}}) := \mathbb{E}_{(\mathcal{A},\mathbf{y})\sim P}\| \op{A}(\widehat{\mathbf{B}})- \mtx{y}\|_2^2 \qquad \text{ s.t. } \qquad \|\widehat{\mathbf{B}}\|_* \leq \lambda.
\end{equation}
Since we have only access to the observed temporal slices of size $d+1$ running up to time $T$, we now build the empirical analog to Program \eqref{eqn:lowrankrecoverynuclear},
\begin{equation} \label{eqn:lowrankrecoverynuclear.ssa}
\min_{\widehat{\mathbf{B}}} \hat{\ell}(\widehat{\mathbf{B}}) := \frac{1}{T-d} \sum_{t=d+1}^{T} \|\mathcal{A}_t(\widehat{\mathbf{B}}) - \mtx{y}_t\|_2^2  \qquad \text{ s.t. } \qquad \|\widehat{\mathbf{B}}\|_* \leq \lambda,
\end{equation}
which is a Lipschitz smooth convex program on the nuclear ball of radius $\lambda$ \cite{lecSP}. Program \eqref{eqn:lowrankrecoverynuclear.ssa} is exactly the same as Program \eqref{eqn:lin.b.lstq} except placed in a matrix recovery context. We aim to recover the optimal $\mtx{B}$ from the samples while accounting for the global structure. When $\lambda$ is arbitrarily large, there is no constraint on $\widehat{\mathbf{B}}$ and Program \eqref{eqn:lowrankrecoverynuclear.ssa} corresponds to fitting each sequence individually with no global information. On the other extreme, forcing $\widehat{\mathbf{B}}$  to be rank one constrains the model for each sequence to be multiples of each other. The intermediate values of $\lambda$ correspond to various trade-offs between learning the common global structure and attention to the individual sequences.

Program \eqref{eqn:lowrankrecoverynuclear.ssa} can be readily cast as a \gls{sdp} solvable via interior point methods \cite{nemirovLMCO}. However, as the size of $\mtx{B}$ may reach into the hundreds of thousands of decision variables, we turn our discussion instead to solutions via efficient first-order proximal algorithms \cite{Combettes2011, 10.1561/2400000003} analogous to those for linear inverse problems \cite{doi:10.1137/080716542}. We will now describe the proximal setup for nuclear norm minimization in the context of time series embedding \cite{Nesterov_Nemirovski_2013}. To illustrate, consider the following \gls{md} procedure
\begin{equation}\label{eqn:md}
    \mathbf{B}_{k+1} = 
    \Prox_{\mathbf{B}_k} (\gamma_k \nabla_{\mathbf{B}_k}[\ell(\mathbf{B}_{k})]),
    \qquad \mathbf{B}_0 \in \{\mathbf{X} \mid \|\mathbf{X}\|_* \leq \lambda\},
\end{equation}
for an appropriately chosen sequence of steps $\{\gamma_k\}$. The solution at step $k$ is given by the aggregate $\tilde{\mathbf{B}}_{k} = (\sum_{\tau=1}^k \gamma_\tau)^{-1} \sum_{\tau=1}^k \gamma_\tau \mathbf{B}_\tau$. We take \textit{prox-mapping} $\Prox_{\mathbf{Z}}(\mathbf{X}) = \argmin_{\|\mathbf{Y}\|_* \leq \lambda} \omega(\mathbf{Y}) + \langle \mathbf{X} - \nabla_{\mathbf{Z}}[\omega(\mathbf{Z})], \mathbf{Y}\rangle$ and \textit{Bregman divergence} $\omega$ to be ones associated with the nuclear ball. In particular, we can compute the nuclear-norm prox mapping by \gls{svt}, eliminating the small singular values at every step \cite{doi:10.1137/080738970, Nesterov_Nemirovski_2013}. Namely, with $m:= \min(d, N)$; $q:= \frac{1}{2 \ln(2m)}$; $\alpha:=\frac{4\sqrt{e} \log(2m)}{2^q(1+q)}; \mathbf{M} = \mathbf{U}\diag\{\sigma_i\}_{i=1}^m\mathbf{V}^*$ we have the Bregman divergence and its subgradient as
\begin{equation}\label{eqn:bregman}
    \omega(\mathbf{M}) = \alpha \sum_{i=1}^m \sigma_i^{1+q} \implies 
    \partial_{\mathbf{M}}[\omega(\mathbf{M})] = \sum_{i=1}^m [\alpha (1+q) \sigma_i^q] \mathbf{u}_i \mathbf{v}_i^*.
\end{equation}
To compute the prox-mapping, consider the \gls{svd} of $\mathbf{X} - \partial_{\mathbf{Z}}[\omega(\mathbf{Z})] = \mathbf{P}\diag\{\sigma_k\}_{k=1}^r\mathbf{Q^*}$. The optimal value of the linear program
\begin{equation} \label{eqn:prox.NN.lp}
    \mathbf{t} = \min_{\mathbf{t} \in \mathbb{R}^n} \{\sum_{i=1}^m\frac{1}{2} t_j^2 - \sigma_i t_i \mid \mathbf{t}\geq 0, \sum_{j=1}^m t_j \leq \lambda\}
\end{equation}
gives $\Prox_{\mathbf{X}}(\mathbf{B}) = \mathbf{U} \diag\{-\mathbf{t}\}_{t=1}^m\mathbf{V}^*$ as the prox-mapping associated with the nuclear ball \cite{Nesterov_Nemirovski_2013}. Notice that Linear Program \eqref{eqn:prox.NN.lp} can be solved in time $\mathcal{O}(m)$ in the worst case. The time for each iteration of Mirror Descent \eqref{eqn:md} is dominated by the cost of the \gls{svd} and the cost to compute the gradient of the loss function.
\subsection{Nonlinear time-series embedding by monotone VI} \label{sec:AR.nonlinear}
We will now extend our discussion to the nonlinear case. Consider again the events $\mathbf{x}_{i,t} \in \mathbb{R}^C$, and allow $\bm{\xi}_{i,t} = \vc(1,\{\mathbf{x}_{i,t-s}\}_{s=1}^d) \in \mathbb{R}^{Cd+1}$ to encode the past $d$ observations with bias term. Allow $\eta: \mathbb{R}^C\to\mathbb{R}^C$ to be a fixed \textit{monotone link function}. Then consider the observation model
\begin{equation}\label{eqn:nonlinearglm}
    \mathbb{E}[\mathbf{x}_{i,t} | \mathcal{H}_{t}] = \eta(\mathbf{R}_{i}\bm{\xi}_{i,t}).
\end{equation}
 Our goal is to form a rank-constrained stochastic estimate to $\mathbf{B}=[\vc(\mathbf{R}_i)]_{i=1}^N \in \mathbb{R}^{(C^2d+ C) \times N}$. However, with arbitrary monotone link function, the \gls{ls} approach outlined in  \eqref{eqn:lin.b.lstq} and \eqref{eqn:lowrankrecoverynuclear.ssa} loses convexity and computational tractability in general. Likewise, \gls{mle} based parameter estimation also becomes computationally difficult \cite{statinferenceconvex}. By contrast, we shall cast the parameter recovery problem into a monotone \gls{vi} formulation, the most general type of convex program where there are known methods to efficiently find high accuracy solutions \cite{juditsky2023generalized, ssso, 9273032}.

\paragraph{Preliminaries on monotone \gls{vi}}
A \textit{monotone vector field} on $\mathbb{R}^m$ with modulus of convexity $\beta$ is a vector field $G: \mathbb{R}^m \to \mathbb{R}^m$ such that
\begin{equation*}
    \langle G(\mathbf{x}) - G(\mathbf{x}^\prime), \mathbf{x}- \mathbf{x}^\prime\rangle \geq \beta \|\mathbf{x} - \mathbf{x}^\prime\| \qquad \forall \ \mathbf{x},\mathbf{x}^\prime \in \mathcal{X}
\end{equation*}
when $\beta > 0$, $G$ is \textit{strongly monotone}. For some convex compact set $\mathcal{X} \subseteq \mathbb{R}^m$, a point $\mathbf{x}^*$ is a \textit{weak solution} to the \gls{vi} associated with $(G, \mathcal{X})$ if for all $\mathbf{x} \in \mathcal{X}$ we have $ \langle G(\mathbf{x}), \mathbf{x} - \mathbf{x}^* \rangle \geq 0$. If $G$ is strongly monotone and a weak solution exists, then the solution is unique. When $\langle G(\mathbf{x}^*), \mathbf{x} - \mathbf{x}^* \rangle \geq 0$ for all $\mathbf{x} \in \mathcal{X}$, we term $\mathbf{x}^*$ a \textit{strong solution} to the \gls{vi}. When $G$ is continuous on $\mathcal{X}$, all strong solutions are weak solutions and vice versa. 

\paragraph{Monotone VI for nonlinear parameter recovery}

We now turn our attention to the construction of a Monotone \gls{vi}, which has as its root optimal parameters corresponding to Model \eqref{eqn:nonlinearglm}. We will use the same operator $\mathcal{A}$ from the linear case from  \eqref{eqn:op.var} together with its associated adjoint $\mathcal{A}^*$. Recall that $\mathcal{A}$ is a random operator that draws upon the concrete time-dependent samples $\mathcal{A}_t$ (which is coupled with observation $\mathbf{y}_t$ of values for the focal time-point across all observed sequences) for some random choice of $t$. The corresponding adjoint $\mathcal{A}_t^*: \mathbb{R}^{CN} \to \mathbb{R}^{(C^2d+C) \times N}$ takes the pre-image of the multichannel predictions (observations) $\mathbf{y} = \vc([\mathbf{x}_{i}]_{i=1}^N)$ and maps them back to the parameter space, and may be computed using the below formula
\begin{equation}
     \mathcal{A}_t^*(\mathbf{y}) = \frac{1}{N} [\vc([x_{i,c} \mathbf{R}_i^T\mathbf{e}_c])_{c=1}^C)]_{i=1}^N: \mathbb{R}^{CN}\to \mathbb{R}^{(C^2d+C) \times N} 
\end{equation}

where $\mathbf{e}_c$ is $c^{\text{th}}$ standard basis. The adjoint, for each entry and channel, multiplies the parameters by the value of the observation corresponding to the channel.

We consider again the accompanying noisy observations $\mathbf{y}$ such that in expectation $\mathbb{E}[\mathbf{y}|\mathcal{A}] = \eta(\mathcal{A}(\mathbf{B}))$ as discussed in Section \ref{sec:AR.linear} and where we extend the link function $\eta$ acts sample wise. Consider now the vector field on the space of matrices
\begin{equation}\label{eqn:MonotoneVI}
    \Psi(\widehat{\mathbf{B}}) = 
    \mathbb{E}_{(\mathcal{A}, \mathbf{y}) \sim P}[ \mathcal{A}^*(\eta(\mathcal{A}(\widehat{\mathbf{B}}))- \mathbf{y})] : \mathbb{R}^{(C^2d+ C) \times N}\to\mathbb{R}^{(C^2d+ C) \times N}
\end{equation}
and notice that the matrix $\mtx{B}$ of true generating parameters is a zero of $\Psi$,
\begin{align*}
    \Psi({\mathbf{B}}) &= 
    \mathbb{E}_{(\mathcal{A}, \mathbf{y}) \sim P}[ \mathcal{A}^*(\eta(\mathcal{A}({\mathbf{B}}))- \mathbf{y})] = \mathbb{E}_{(\mathcal{A}, \mathbf{y}) \sim P} [\mathcal{A}^*(\eta(\mathcal{A}({\mathbf{B}})))-\mathcal{A}^*(\mathbf{y})]\\
    &=\mathbb{E}_{(\mathcal{A}, \mathbf{y}) \sim P} [\mathcal{A}^*(\eta(\mathcal{A}({\mathbf{B}})))-\mathcal{A}^*(\mathbb{E}[{\mathbf{y} \mid \mathcal{A}}])]\\
    &= \mathbb{E}_{(\mathcal{A}, \mathbf{y}) \sim P} [\mathcal{A}^*(\eta(\mathcal{A}({\mathbf{B}})))-\mathcal{A}^*(\eta(\mathcal{A}({\mathbf{B}})))]=0
    .\end{align*}
Since we have only access to the given observations, we take solutions 
to the empirical version of the \gls{vi} that takes slices from the time-series 
\begin{equation}
\label{eqn:vi.ssa}
\widehat{\Psi}(\widehat{\mathbf{B}}) = \frac{1}{T-d}\sum_{t=d}^{T} [\mathcal{A}_t^*(\eta(\mathcal{A}_t(\widehat{\mathbf{B}}))) - \mathcal{A}_k^*(y_k)] = \frac{1}{T-d}\sum_{t=d+1}^{T} \mathcal{A}_t^*[ \eta(\mathcal{A}_t(\widehat{\mathbf{B}})) - \mathbf{y}_t].
\end{equation}
At each time window $t$, $\mathbf{y}_t$ represents the sequence observations. $\mathcal{A}_t$ then takes in as input an estimate to matrix $\mathbf{B}$ and uses the data from the previous $d$ time-points to output a prediction for $\mathbf{y}_t$. We note that both $\mathcal{A}_t$ and $\mathcal{A}_t^*$ may be computed in time $\mathcal{O}(NC^2d)$. In most of our computations, we form an approximation to $\mathcal{A}$ by averaging across the entire time horizon, giving a cost of $\mathcal{O}(TNC^2d)$. We also illustrate averaging using smaller random sub-windows of the data in Section \ref{sec:exp.gene}. Analogous to Program \eqref{eqn:lowrankrecoverynuclear.ssa}, the \gls{vi} associated with  \eqref{eqn:vi.ssa} likewise admits solutions by \gls{md}. To illustrate, consider the recurrence:
\begin{equation} \label{eqn:MD.VI}
    {\mathbf{B}}_{k+1} = \Prox_{{\mathbf{B}}_k}(\gamma_k \Psi(\widehat{\mathbf{B}}_k)) \qquad \mathbf{B}_0 \in \{{\mathbf{X}} \mid \|{\mathbf{X}}\|_* \leq \lambda\}
\end{equation}
with step sizes $\{\gamma_k\}$. The aggregate solution at step $k$ is given by $\tilde{\mathbf{B}}_{k} = (\sum_{\tau=1}^k \gamma_\tau)^{-1} \sum_{\tau=1}^t \gamma_\tau {\mathbf{B}}_\tau$ \cite{doi:10.1137/S1052623403425629}. Note if $\eta := \mathbf{Id}$, the identity function, then the \gls{vi} corresponds exactly to the \textit{gradient field} of Program \eqref{eqn:lowrankrecoverynuclear.ssa}. In this case,  
$\Psi({\mathbf{X}}) = \nabla_{\mathbf{X}}[\ell(\mathbf{X})]$ and the \gls{md} procedure for \gls{vi} and \gls{ls} are the same.

\paragraph{First order methods for monotone VI}
To concretely solve the monotone \glspl{vi} outlined in  \eqref{eqn:MonotoneVI} and \eqref{eqn:vi.ssa},  we detail an accelerated mirror-prox scheme with backtracking for nuclear norm constrained \gls{vi} in Algorithm \ref{alg:mp} of Appendix \ref{apx:impl},  which addresses the following general problem
\begin{equation}
    \langle \Psi (\mathbf{B}), \mathbf{B}- \mathbf{B}^* \rangle \geq 0 \qquad \forall \mathbf{B} \in \mathcal{X}:=\{\mathbf{B} \mid \|\mathbf{B}\|_* \leq \lambda\}
\end{equation}
where $\Psi$ is an (unbiased estimator of) $\kappa$-lipschitz continuous monotone vector field  \cite{Chen:2017aa,Nesterov_Nemirovski_2013}, which addresses the difficulty that $\kappa$ is in most cases is unknown to us beforehand. The convergence results for this class of algorithm are typical and presented in \cite{Nesterov_Nemirovski_2013, Chen:2017aa, doi:10.1137/S1052623403425629, doi:10.1287/10-SSY011}. Namely, for $\epsilon$ error as measured by $\epsilon_{\Psi, \mathcal{X}}(\tilde{\mathbf{B}}_t) = \sup_{\mathbf{Z} \in \mathcal{X}} \langle \Psi(\mathbf{Z}, \tilde{\mathbf{B}}_t - \mathbf{Z})\rangle$ solution requires $\mathcal{O}({\kappa}/{\epsilon})$ iterations for deterministic \glspl{vi} and $\mathcal{O}(\kappa/{\epsilon}+ \sigma^2/{\epsilon}^2)$ for $\mathbb{E}[\|\widehat{\Psi}(\mathbf{B})- \Psi(\mathbf{B})\|^2]\leq \sigma^2$ where $\widehat{\Psi}(\mathbf{B})$ is a stochastic approximation to the true field $\Psi$ with bounded variance. Each iteration requires $\mathcal{O}(1)$ evaluations of the monotone operator $\Psi$ and $\Prox$ operator, which in turn is dominated by the cost of the \gls{svd} at each step $\mathcal{O}(NC^2d \min(N, C^2d))$, with the trade-off being in the variance of the approximation to the true VI field. The convergence of the algorithm as applied to the vector fields given in  \eqref{eqn:MonotoneVI} and \eqref{eqn:vi.ssa} in particular may be established similarly to \cite{ssso}, and when the data follow the true model \eqref{eqn:nonlinearglm}, parameter recovery guarantee can be established similarly to \cite{9273032}.

\paragraph{Parameter recovery for symbolic sequences}
\label{sec:SEQ}
As a special case of \eqref{eqn:MonotoneVI}, consider now that each channel represents the probability of emitting a token taken from syllabary $\{s_c\}_{c=1}^C$ of size $C$. Then each $\mathbf{x}_{i,t}$ represents a probability vector 
$\sum_{c}^C x_{i,t,c}=1$ and we have $\mathbb{E}[x_{i,t,c} | \mathcal{H}_{t}] = \mathbb{P}[x_{i,t,c}=s_c]$. We take the softmax activation function $\sigma(\mathbf{y}) = \vc([\|\exp(\mathbf{y}^{(i)})\|_1^{-1} \exp(\mathbf{y}^{(i)})]_{i=1}^N)$, where $\mathbf{y}^{(i)}$ corresponds to values from the $i^\mathrm{th}$ sequence. This problem corresponds to learning representations for different sequences. We illustrate in Section \ref{sec:exp.vigngettes} the above as applied to learning dynamics for genomics data and natural language, for which autoregressive models have become increasingly popular.

\section{Experiments}
\label{sec:exp}
We first illustrate parameter recovery using synthetic univariate time-series in Section \ref{sec:exp.syn}. We investigate the choice of nuclear penalty $\lambda$ and the geometry of the recovered parameter matrix. Section \ref{sec:exp.UCRreal} describes benchmarks using real-world time-series data from the UCR Time Series Classification Archive \cite{UCRArchive2018}. We report classification and runtime performance against a number of baselines. Section \ref{sec:exp.vigngettes} provides two illustrations on the embedding of real-world sequence data. In the first example, we consider a language representation task where we embed without supervision a series of excerpts taken either from the works of Lewis Carroll or abstracts scraped from arXiv \cite{alice1,alice2, arxiv-dataset}.  In the second illustration, we apply our method to the clustering of gene sequences for strains of the \textit{Influenza A} and \textit{Dengue} viruses \cite{Sayers2022-zf}. 

We evaluated all experiments and illustrations using a cluster with 24 core Intel Xeon Gold 6226 CPU (2.7 GHZ) processors, and NVIDIA Tesla V100 Graphics coprocessors (16 GB VRAM), and 384 GB of RAM. However, we could reproduce the results on a standard personal computer. We implement the routines described in Sections \ref{sec:AR.linear} and \ref{sec:AR.nonlinear} and Appendix \ref{apx:impl} in the Julia programming language. We describe the experimental setup and results in detail in Appendix \ref{apx:exp}. 

\subsection{Parameter recovery with synthetic autoregressive sequences}
\label{sec:exp.syn}
To illustrate parameter recovery across autoregressive sequences, we synthetically generated a set of ten baseline parameters for linear autoregressive sequences of order $d=15$. Within each class, we then created $N=300$ sequences of each type, perturbing the baseline coefficients by adding a small amount of noise according to a fixed rule for the set of parameters. We then generated $T=250$ observations for each sequence according to the autoregressive model in  \eqref{eqn:univar.reg}. We formed all 120 combinations of $k=3$ type of sequences from the ten classes and recovered the underlying parameter matrix by solving the program given in  \eqref{eqn:lowrankrecoverynuclear.ssa} to optimality. We report the data-generating procedure and experimental details in Appendix \ref{apx:exp.syn}.

Table \ref{tab:synthetic} reports averages and standard deviations for the relative reconstruction error $\|\mathbf{B}-\widehat{\mathbf{B}}\|_F/\|\mathbf{B}\|_F$, the least squares error of the objective function given in  \eqref{eqn:lin.b.lstq}, the \gls{ari} using k-means clustering with $k=3$ clusters in the embedding space across the runs, and the number of large singular values (singular values within $10^{-2}$ of the principal singular value). We first learn representations for sequences without nuclear norm constraint.

To illustrate the low-rank matrix recovery, we search across the values of nuclear regularization $\lambda$ via Brent search \cite{brent} and report the performance for a close to the optimal value of $\lambda$ with respect to the reconstruction error. When the underlying dynamics share a common low-rank structure, the nuclear constraint effectively leverages the common information shared across the sequences to recover the true parameters more faithfully using the common domain information with comparable error in the least squares sense. Furthermore, we observe the nuclear regularization procedure driving the singular value spectrum to sparsity, with the number of large singular values being much smaller than in the unconstrained case.

\begin{figure}
  \centering
    \includegraphics[width=\textwidth]{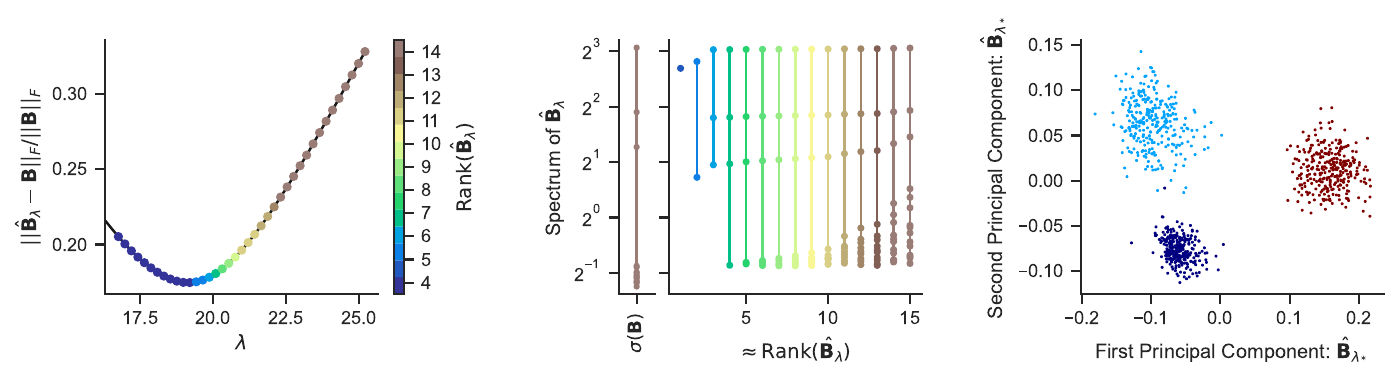}
  \caption{Simulation results: Parameter recovery for a collection of univariate time series drawn from $k=3$ classes. Left: Relative reconstruction error and approximate rank of recovered parameter matrices across levels of nuclear constraint $\lambda$. Center: Singular values of the true parameter matrix $\mathbf{B}$ and the singular values of the recovered solutions of varying dimensions. Right: First two principal components of the recovered matrix with the smallest reconstruction error and original class labels.}
  \label{fig:synthetic}
\end{figure}

To further illustrate, Figure \ref{fig:synthetic} depicts parameter recovery for a collection of sequences with autoregressive order $d=15$. In the leftmost pane, we report the relative reconstruction error and the number of large singular values across different values of nuclear constraint $\lambda$. In the central pane, we depict the singular value spectra of the true parameter matrix (which has approximately low-rank structure plus noise) and recovered matrices with differing numbers of large singular values. In the third pane, we show the first two principal components (total explained variance $=0.868$) of the sequence embedding with the smallest reconstruction error \cite{pml1Book} along with the original three generating classes of the data. With the introduction of sufficient nuclear regularization, we drastically reduced the reconstruction error of our recovered solution and observed that the solutions with low reconstruction error were of approximately low rank. In the plots of the singular value spectra and the projection of the learned sequence embeddings, we observe that those rank-constrained recoveries effectively recover those large singular values in the spectrum of the true parameter matrix. By contrast, the parameter recoveries performed without or insufficient nuclear constraint fit the noise component of the data, as evident in the distribution of singular values.

 \begin{table}
   \caption{Time series parameter recovery for synthetic autoregressive time-series.}
   \label{tab:synthetic}
   \centering
   \small
   \begin{tabular}{ccccccccc}
    \toprule
     
       & \multicolumn{2}{c}{Relative Err.} & \multicolumn{2}{c}{LS Err.} & \multicolumn{2}{c}{Cluster ARI \cite{ari}} & \multicolumn{2}{c}{$\approx$ Rank}\\
       \cmidrule(r){2-3} \cmidrule(lr){4-5} \cmidrule(lr){6-7} \cmidrule(l){8-9}
     $\lambda$ Selection & Avg. & Std.& Avg. & Std.& Avg. & Std.& Avg. & Std.\\
     \midrule
     Unconstrained &0.341 &(0.016) & 79.333 & (0.169)  & 0.967 & (0.049) & 14.442 & (1.203)    \\
     $\argmin_\lambda \|\widehat{\mathbf{B}}_\lambda- \mathbf{B}\|_F$ &0.158 & (0.020) & 80.709 & (0.233) & 0.997 & (0.008) & 7.392 & (4.255)\\
     \bottomrule
   \end{tabular}
 \end{table}

\subsection{Real time-series classification}
\label{sec:exp.UCRreal}
Following in the line of \cite{Middlehurst, NEURIPS2019, Yue_Wang_Duan_Yang_Huang_Tong_Xu_2022, shaplets2}, we conduct real data experiments using 36 of the UCR time series datasets \cite{UCRArchive2018}. We report  statistics for the time series datasets in Appendix \ref{apx:exp.ucr.stats}. Each dataset has a default test/train split. We first re-encoded each of the time series as a multichannel signal consisting of the signal itself and its first finite-difference. We then embed all the data without supervision by solving  \eqref{eqn:vi.ssa} using the adaptive mirror prox scheme described in Algorithm \ref{alg:mp} of Appendix \ref{apx:impl} with a look-back length of $d=20$ and running the algorithm for $256$ steps, taking the link function to be linear. To pick the value of $\lambda$, we find a value when the solution becomes rank one via bisection and perform a grid search across values of $\lambda$, which results in rank-constrained parameters. We then report our results using the choice of $\lambda$ with the best performance on the training set. We report the \gls{ari} \cite{ari}, \gls{nmi} \cite{10.1145/1553374.1553511}, macro-F1 score \cite{FAWCETT2006861}, and accuracy on the test set. We also report the average runtime.

We compare our method with five representative time series embedding and classification methods: \gls{knn} using Euclidean ($\ell_2$) distance \cite{1053964}, \gls{knn} with \gls{dtw} as the distance metric \cite{Muller2007}, shapeDTW (another method based on \gls{dtw} but with additional features) \cite{shaplets2, 10.1145/1557019.1557122}, a dictionary-based method MultiROCKET+Hydra \cite{Dempster:2023aa}, and one deep-learning representation method based on contrastive learning (TS2Vec) \cite{Yue_Wang_Duan_Yang_Huang_Tong_Xu_2022}. In line with \cite{Yue_Wang_Duan_Yang_Huang_Tong_Xu_2022, NEURIPS2019_53c6de78}, to evaluate the classification performance on test set for methods which produce embeddings (TS2Vec and our method),  we perform cross-validated grid search across \glspl{knn} with $k=\{2^i \mid i \in[0,4]\}$ neighbors or SVMs with RBF kernels with penalty values $c\in\{2^i \mid i \in[-10,15]\}\cup{\infty}$. We defer all further details of our experimental setup to Appendix \ref{apx:exp}. Table \ref{tab:ucr} displays the mean and standard deviation across the metrics across the datasets. We provide the detailed results per dataset in Tables \ref{tab:resultbyDataset1} and \ref{tab:resultbyDataset2} of Appendix \ref{apx:exp.ucr}. We observe superior performance to baseline methods based on distance metrics, such as Euclidean distance or \gls{dtw}, and observe comparable performance to TS2Vec. We note that for this class of univariate sequence, the heuristic dictionary-based ensemble (MR-HYDRA) outperforms both our approach and the deep-learning-based approaches. However, this method has been tuned specifically for this type of classification problem. By contrast, similar to \cite{Yue_Wang_Duan_Yang_Huang_Tong_Xu_2022}, we consider classification only as one potential downstream task and do not rely on handcrafted features.

 \begin{table}
   \caption{Time series classification performance on UCR time series data of our method vs a number of baselines (higher is better, except for runtime). We outperform simple approaches and perform close to classification using the embeddings from the neural network based TS2Vec but use only 37\% of the runtime. The best performing method, MR-Hydra, is a ensemble based on handpicked features tuned specially for time series classification. }

   \label{tab:ucr}
   \centering
   \scriptsize
   \begin{tabular}{lllllllllll}
     \toprule
               & \multicolumn{2}{c}{ARI \cite{ari}} & \multicolumn{2}{c}{NMI \cite{10.1145/1553374.1553511}} & \multicolumn{2}{c}{F1\cite{FAWCETT2006861}} & \multicolumn{2}{c}{Accuracy} & \multicolumn{2}{c}{Runtime (Sec)}\\
               \cmidrule(r){2-3} \cmidrule(lr){4-5}  \cmidrule(lr){6-7} \cmidrule(l){8-9} \cmidrule(l){10-11} 
     Method    & Avg. & Std. & Avg. & Std. &Avg. & Std.& Avg. & Std. & Avg. & Std.
     \\\midrule
     $\ell_2$+KNN \cite{1053964} &0.422 &(0.294)& 0.416 &(0.290)& 0.752 &(0.144)& 0.725 &(0.166)& 0.128&(0.413) \\
     DTW+KNN \cite{Muller2007}& 0.447 &(0.303)& 0.435 &(0.300)& 0.766 &(0.150)& 0.738 &(0.170)& 42.988&(134.320)\\
     shapeDTW \cite{ZHAO2018171} & 0.470 &(0.307)& 0.460 &(0.299)& 0.773 &(0.152)& 0.746 &(0.178)& 21.871&(71.655) \\     
     % Inception Time \cite{IsmailFawaz2020inceptionTime} & 0.461 &(0.356)& 0.471 &(0.342)& 0.738 &(0.210)& 0.681 &(0.257)& 243.226&(185.928) \\
     TS2Vec \cite{Yue_Wang_Duan_Yang_Huang_Tong_Xu_2022}& 0.606 &(0.282)& 0.580 &(0.287)& 0.840 &(0.138)& 0.814 &(0.178)& 1,085.092&(1,408.458)\\
     Ours& 0.602 & (0.282) & 0.562 & (0.293) & 0.817 & (0.180) & 0.788 & (0.193) & 400.031& (677.486) \\
     \midrule MR-Hydra \cite{Dempster:2023aa} & {0.682} &(0.273)& {0.656} &(0.285)& {0.877} &(0.121)& {0.851} &(0.162)& 10.197&(16.459)\\
     \bottomrule
   \end{tabular}
 \end{table}

\subsection{Symbolic sequences: language and genomics}
\label{sec:exp.vigngettes}
\begin{figure}
     \centering
     \begin{subfigure}[b]{0.32\textwidth}
         \centering
         \includegraphics[width=\textwidth]{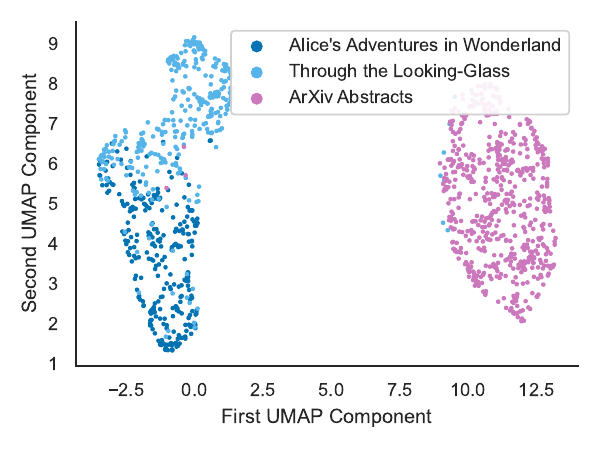}
         \caption{Lewis Caroll or ArXiv abstracts? Genres form clusters.}
         \label{fig:subfiga}
     \end{subfigure}
     \begin{subfigure}[b]{0.32\textwidth}
         \centering
         \includegraphics[width=\textwidth]{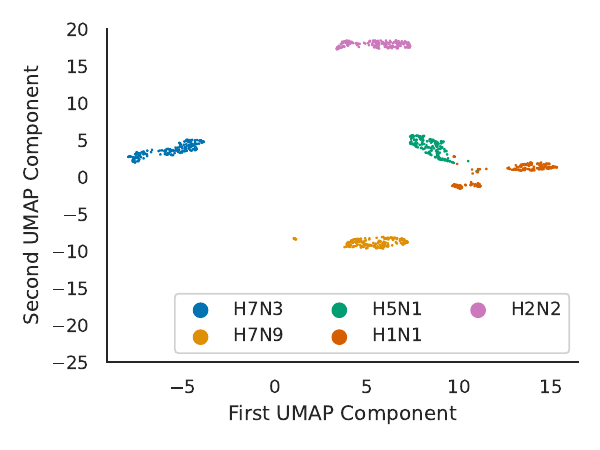}
         \caption{Clustering strains of \textit{Influenza A virus} genome data (segment 6)}
         \label{fig:subfigb}
     \end{subfigure}
     \begin{subfigure}[b]{0.32\textwidth}
         \centering
         \includegraphics[width=\textwidth]{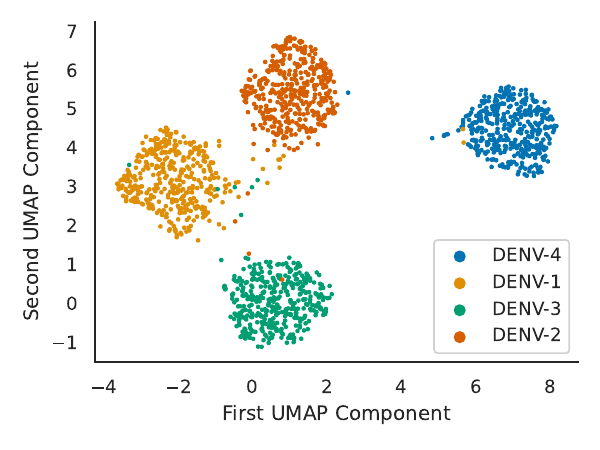}
         \caption{Clustering using full genome for the four strains of \textit{Dengue Virus}}
         \label{fig:subfigc}
     \end{subfigure}
     \caption{Learned embeddings for symbolic sequences collections using our method--- visualised by UMAP projections shows clear groupings based on sequences with similar underlying dynamics.}
     \label{fig:symb}
 \end{figure}

We now illustrate the capability of our method to learn meaningful representations for sequences with nonlinear dynamics, namely autoregressive sequence and language modeling.
\paragraph{Symbolic sequences and language: arXiv abstracts or ``Alice in Wonderland''? }
\label{sec:exp.language}
We first consider an autoregressive language modeling task, drawing textual sequences from three sources: two works by the same author Lewis Caroll ---  \textit{Alice's Adventures in Wonderland} ($n=228$) \cite{alice1} and \textit{Through the Looking Glass} ($n=316$) \cite{alice2} --- and machine learning related abstracts scraped from ArXiv ($n=600$)  \cite{arxiv-dataset}  (details may be found in Appendix \ref{apx:exp.lang}). We embed the sequences without supervision with a lookback of $d=75$, and in order to reduce the number of symbols in our alphabet and avoid the blowup in the number of channels, we converted each of the sequences into a $c=4$ symbol code via Huffman coding, based on the overall frequencies of letters in the English language \cite{huffman}. We then solve Program \eqref{eqn:vi.ssa} using the multichannel measurement operator given in  \eqref{eqn:op.var} to optimality and using the softmax activation discussed in Section \ref{sec:SEQ}. We show in Figure \ref{fig:subfiga} the space learned when $\lambda$ was chosen to be sufficiently small as to give a rank three representation of the data. We then project the learned representation via UMAP \cite{McInnes2018}. Two distinct clusters form corresponding to the two different genres of writing, however, whereas the paper abstracts are clearly separable from the works of Lewis Caroll, the two books written by him are not as clearly disambiguable considering they are from the same author.

\paragraph{Virus strain identification from genome sequences}
\label{sec:exp.gene}
For the final illustration, in line with \cite{10.1371/journal.pone.0261531}, the problem of classifying genetic sequences, which allows for the placing of species/variants in the evolutionary context of others (phylogenetic classification). We consider gene sequence data from segment six of the \textit{Influenza A virus} genome ($n=949$, $\text{average sequence length}=1409$) \cite{Bao2008-oq} and the complete genome the \textit{Dengue virus} ($n=1633$, $\text{average sequence length}=10559$) \cite{Hatcher2017-hh}. We consider gene sequences from five strains of Influenza and the four strains of Dengue. Likewise, we provide a detailed overview of the data and learning procedure in Appendix \ref{apx:gene}. We encode the genomes in a similar manner as for the natural language illustration, assigning one channel to each nucleotide (A, C, T, G), and encode the presence/absence of each nucleotide at each position via one-hot encoding. 

To recover the embedding, we adopt the same softmax activation scheme as described in Section \ref{sec:SEQ}. Since the genomes are of variable length, we consider a stochastic approximation to the monotone field $\Psi$  \eqref{eqn:MonotoneVI} by taking the sample average of randomly selected length $G=800$ sub-windows from each of sequences at each training step. We consider clustering the Influenza and Dengue genome segments individually and report UMAP projections of the learned representations in Figure \ref{fig:subfigb} and \ref{fig:subfigc}, respectively. The dimensions of the learned embeddings are 7 and 20, respectively. In these subspaces, we note the clear grouping of viral strains obtained via solving the stochastic approximation to the VI \eqref{eqn:MonotoneVI}.

\section{Discussion}

 We have introduced a method to learn low-dimensional representations for collections of nonlinear time series with provable recovery guarantees. We framed the problem as a low-rank matrix recovery problem that is representable as a monotone variational inequality, the most general type of convex problem for which efficient solution methods exist. We suppose the sequence dynamics obey an autoregressive model via a monotone link function but make no further assumptions about the sequences. This notably includes the important case of the probabilistic modeling of symbolic sequences. Our method exhibits strong performance when these assumptions are satisfied. However, the method is limited when the data falls short of these assumptions, as is the case for some UCR time-series data in Section \ref{sec:exp.UCRreal}. To address this, we note that we may also design alternative objectives as long they are representable in the form of monotone VI. The runtime performance of our method is dominated by (1) the cost of the \gls{svd} to enforce the low-rank constraint and (2) computing a high-quality estimate to the underlying VI by sample stochastic approximation. However, as shown in the Section \ref{sec:exp.gene} genome representation example, we can take random windows substantially shorter than the entire sequence to estimate the VI field and still maintain competitive performance. Furthermore, if the low-rank assumption is satisfied and the majority of the signal energies are concentrated in the principal singular values, at each $\Prox$ (thresholding) step, we need only compute singular values up to $\lambda$. In this setting, a good idea is to apply an iterative Lanczos method to find the first few singular values/vectors in an on-line fashion \cite{doi:10.1137/080738970, doi:10.1137/1.9781611977165, doi:10.1137/1.9781421407944}. However, we discuss neither these acceleration methods nor loosening the class of the model while still maintaining convexity in the scope of this work. 

\printbibliography

\appendix

\section{Implementation Details}
\label{apx:impl}

\subsection{First order methods for monotone VI with nuclear ball setup}
\label{apx:nbfom}

We present the concrete accelerated mirror-prox method with backtracking for nuclear norm constrained \gls{vi} based on \cite{Chen:2017aa,Nesterov_Nemirovski_2013}, in Algorithm \ref{alg:mp}.

\begin{algorithm}
    \caption{Accelerated Mirror Prox Method with Backtracking for Nuclear Norm constrained \gls{vi}}
    \label{alg:mp}
    \begin{algorithmic}[1]
    \Procedure{AcceleratedBacktrackingMirrorProx}{$\Psi$, $N$, $\lambda$, $\kappa_0$}

    \Comment{$\Psi$ monotone \gls{vi}, $N>0$ number of steps, $\lambda>0$ radius of nuclear ball, $\kappa_0>0$ initial guess of Lipschitz constant}
    \State $\mathbf{R}_1  := \mathbf{0}, \mathbf{B}_1 := \mathbf{R}_1, \tilde{\mathbf{B}}_1 := \mathbf{R}_1$
    
    \For{$t:= 1, N$}
    \State $\hat{\kappa}_t := \kappa_{t-1}$
    \Repeat
    \State $\alpha_t := \frac{2}{t+1}, \hat{\gamma}_t := \frac{t}{2\hat{\kappa}_t t}$
    \State $\mathbf{B}_{t+1} := \text{\sc{ProxNuc}}(\mathbf{R}_t, \hat{\gamma}_t \Psi(\mathbf{R}_t))$
    \If{$\|\Psi(\mathbf{B}_{t+1})-\Psi(\mathbf{R}_{t})\|_F > \hat{\kappa}_t\|\mathbf{B}_{t+1}-\mathbf{R}_t\|_F$}
    \State $\hat{\kappa}_t := 2 \hat{\kappa}_t$
    \EndIf
    \Until{$\|\Psi(\mathbf{B}_{t+1})-\Psi(\mathbf{R}_{t})\|_F \leq \hat{\kappa}_t\|\mathbf{B}_{t+1}-\mathbf{R}_t\|_F$} 
    \State $\kappa_t := \hat{\kappa}_t$
    \State $\gamma_t = \hat{\gamma}_t$
    \State $\mathbf{R}_{t+1} := \text{\sc{ProxNuc}}(\mathbf{R}_t, {\gamma}_t \Psi(\mathbf{B}_{t+1}))$
    \State $\tilde{\mathbf{B}}_{t+1} := (1-\alpha_t)\tilde{\mathbf{B}}_{t} + \alpha_t \mathbf{B}_{t+1}$
    \State \textbf{return} $\tilde{\mathbf{B}}_{N+1}$
    \EndFor
    \EndProcedure
    \Procedure{ProxNuc}{$\mathbf{Z}$,$\mathbf{X}, \lambda$}\Comment{Computes $\Prox_{\mathbf{Z}, \|\cdot\|_{*}\leq\lambda}(\mathbf{X})$, $\mathbf{Z} \in \mathbb{R}^{m\times n}$}
    \State $r:= \min(m,n)$
    \State $q:= (2 \log 2r)^{-1}$
    \State $\mathbf{Y} := x-\partial \omega(\mathbf{Z})$
    \State $\mathbf{U},\bm{\delta}, \mathbf{V}^* := \text{\sc{svd}}(\mathbf{Y})$
    \State $\mathbf{t} := \min_{\mathbf{t} \in \mathbb{R}^r} \{\frac{1}{2} \sum_{j=1}^{m} t_j^2 - \delta_j t_j \mid t \geq \mathbf{0}, \sum_{j=1}^m t_j \leq \lambda\}$
    \State \textbf{return} $\mathbf{U}\diag(-\mathbf{t})\mathbf{V}^*$ 
    \EndProcedure
    \Procedure{$\partial \omega$}{$\mathbf{Z}$} \Comment{Compute subgradient of \acrshort{dgf} $\omega$ from  \eqref{eqn:bregman}, $\mathbf{Z} \in \mathbb{R}^{m\times n}$}
    \State $r:= \min(m,n)$
    \State $q:= (2 \log 2r)^{-1}$
    \State $c:=\frac{4 \sqrt{e} \log(2r)}{2^q(1+q)}$
    \State $\mathbf{U}, \bm{\sigma}, \mathbf{V}^* := \text{\sc{svd}}(\mathbf{Y})$
    \State \textbf{return} $c \sum_i^r (1+q) (\sigma_i)^q \mathbf{u}_i \mathbf{v}^*$
    \EndProcedure
\end{algorithmic}
\end{algorithm}

\section{Detailed Experimental Setup and Results}
\label{apx:exp}

\subsection{Synthetic sequences}
\label{apx:exp.syn}

\subsubsection{Data Generation}
For the synthetic sequence recovery experiment, we adopt the following data-generating procedure: We take the order of the sequences to be $d=15$, and we generate data according to the following procedure within each of the five generated classes of observations
\begin{enumerate}
    \item Pick a baseline set of coefficients according to a given random distribution
    \item For each of the $N=300$ sequences to generate, perturb the coefficients according to the pre-specified rule
    \item Generate the data matrix of size consisting of $T=250$ of the $N=300$ sequences according to the perturbed coefficients such that the data obeys  \eqref{eqn:univar.reg}. To do so, we seat the first 15 observations using random noise such that $x_{i,t}\sim \mathcal{N}(\mu= 0, \sigma^2=1), \forall t\in [1,d], i \in [N]$. Then each successive entry $x_{i,t},  t \in [d+1,T]$ is then given by taking $x_{i,t} = \sum_{s=1}^d{b}_{i,s}x_{i,t-s}+ \epsilon_{i,t}$, $\epsilon_{i,t} \sim \mathcal{N}(\mu=0, \sigma^2=0.02)$.
\end{enumerate}

We draw the ten generated classes of data from the following five generation procedures given in Table \ref{tab:syn.datagen}. We use each procedure twice to generate the ten classes of data. We denote the coefficients common to the sequences (for some class) as $\mathbf{b}_{\text{common}}$, and the coefficients for the $i^{\text{th}}$ sequence in said class as $\mathbf{b}_i$.

The baseline coefficients generation methods are given as:
\begin{description}
    \item[Exponentially Time Decaying:] $b_{\mathrm{common}, s} = Z  \gamma^{s}/(\sum_{j=1}^d \gamma^j) \qquad Z \sim \mathrm{Uniform}([0,1]), \forall s \in [d]$
    \item[Uniform:] $b_{\mathrm{common},s} = Z \qquad Z \sim \mathrm{Uniform}([0,1/2d]), \forall s \in [d]$
\end{description}

and the perturbation methods are given as:
\begin{description}
    \item[Gaussian:] $\mathbf{b}_i = \mathbf{b}_{\mathrm{common}} + \mathbf{Z} \qquad Z_j \sim \mathcal{N}(\mu=0, \sigma^2 = 0.02)$
    \item[$d/3$ Most Recent:] $\mathbf{b}_i = \mathbf{b}_{\mathrm{common}} + \mathbf{Z} \qquad Z_j \sim \begin{cases}\mathcal{N}(\mu=0, \sigma^2 = 0.02) & j< \lceil d/3 \rceil \\ 0\end{cases}, \forall i \in [N]$
    \item[Uniform $\times$ Fixed Vector:]  $\mathbf{b}_i = \mathbf{b}_{\mathrm{common}} + \theta\mathbf{v} \qquad \theta \sim \mathrm{Uniform}([-1,1]), \|\mathbf{v}\| = 1 , \forall i \in [N]$ \\($\mathbf{v}$ chosen uniformly on a unit hypersphere, and is the same for all sequences generated in the class)
\end{description}

\begin{table}
    \centering
    \caption{Five classes of sequence generating procedure}
    \begin{tabular}{ll}
        \toprule
        {Baseline Coefficients} & {Perturbation Pattern} \\
        % \cmidrule(lr){1-2} \cmidrule(lr){3-4}
        % Description & Formula & Description & Formula \\
        \midrule
        Exponentially Time Decaying & Gaussian \\
        Exponentially Time Decaying & $d/3$ Most Recent  \\
        Exponentially Time Decaying  & Uniform $\times$ Fixed Vector  \\
        Uniform & Gaussian  \\
        Uniform & Uniform $\times$ Fixed Vector\\
        \bottomrule
    \end{tabular}
    \label{tab:syn.datagen}
\end{table}

\subsubsection{Parameter Recovery}
For the parameter recovery experiment, we take all $\binom{10}{3}=120$ combinations of $k=3$ sequences from the 10 classes and concatenate the generated sequences to form a matrix of $900$ observations. We then recover the baseline coefficient matrix $\mathbf{B} \in \mathbb{R}^{15 \times 900}$. To recover the parameters for each sequence, we solve the program given in  \eqref{eqn:lowrankrecoverynuclear.ssa} to optimality for differing levels of $\lambda$ (using a standard convex solver by contrast to Algorithm \ref{alg:mp}). To find which levels of $\lambda$ to solve for, we first solve the unconstrained version of the problem. We then compute the nuclear norm of recovered $\|\widehat{\mathbf{B}}\|_*$. We then successively search for the optimal $\lambda^*$ on the interval $[0, \|\widehat{\mathbf{B}}\|_*]$ using the relative reconstruction error $\|\mathbf{B}-\widehat{\mathbf{B}}\|_F/\|\mathbf{B}\|_F$ as the objective until we achieve an absolute tolerance of $10^{-3}$. We report the results in Table \ref{tab:synthetic} using the matrices $\widehat{\mathbf{B}}$ and $\widehat{\mathbf{B}}_{\lambda^*}$ across the 120 runs. For Figure \ref{fig:synthetic}, 
we use data drawn from the classes uniform baseline coefficients with Gaussian perturbation, exponentially time-decaying baseline coefficients with Gaussian perturbation; and exponentially time-decaying baseline coefficients with uniform*fixed vector perturbation. We find the $\lambda^*$ via Brent search, and we sweep 40 values of $\lambda \in [16.3, 25.2]$ (the right bound corresponding to the value of $ \|\widehat{\mathbf{B}}\|_*$) to produce Figure \ref{fig:synthetic}. When reporting the spectra of singular values, we pick solutions that correspond to the largest value of $\lambda$ we have for some fixed number of large singular values. Finally, we depict the principal components according to the recovered matrix $\widehat{\mathbf{B}}_{\lambda^*}$, which is optimal in the sense of the reconstruction error (found via Brent search).

\subsection{UCR time series}
\label{apx:exp.ucr}
\subsubsection{Data Overview}
We compare our method with the following representative time series clustering methods.
\label{apx:exp.ucr.stats}
In line with \cite{NEURIPS2019} we selected 36 of the univariate UCR time series classification datasets \footnote{\url{https://www.cs.ucr.edu/\%7Eeamonn/time_series_data_2018/}}. We report basic statistics (training samples, testing samples, length, and number of classes) of the datasets in Table \ref{tab:sampleData}.

\begin{table}
    \centering
    \caption{Basic statistics for the UCR time-series classification benchmark data}
    \tiny
\begin{tabular}{lllll}
\toprule
 & Training Samples & Testing Samples & Length & Classes \\
\midrule
ArrowHead & 36 & 175 & 251 & 3 \\
Beef & 30 & 30 & 470 & 5 \\
BeetleFly & 20 & 20 & 512 & 2 \\
BirdChicken & 20 & 20 & 512 & 2 \\
Car & 60 & 60 & 577 & 4 \\
ChlorineConc. & 467 & 3840 & 166 & 3 \\
Coffee & 28 & 28 & 286 & 2 \\
DiatomsizeReduction & 16 & 306 & 345 & 4 \\
Dist.Pha.Outln.AgeGrp. & 400 & 139 & 80 & 3 \\
Dist.Pha.Outln.Correct & 600 & 276 & 80 & 2 \\
ECG200 & 100 & 100 & 96 & 2 \\
ECGFiveDays & 23 & 861 & 136 & 2 \\
GunPoint & 50 & 150 & 150 & 2 \\
Ham & 109 & 105 & 431 & 2 \\
Herring & 64 & 64 & 512 & 2 \\
Lightning2 & 60 & 61 & 637 & 2 \\
Meat & 60 & 60 & 448 & 3 \\
Mid.Pha.Outln.AgeGrp. & 400 & 154 & 80 & 3 \\
Mid.Pha.Outln.Correct & 600 & 291 & 80 & 2 \\
Mid.PhalanxTW & 399 & 154 & 80 & 6 \\
MoteStrain & 20 & 1252 & 84 & 2 \\
OSULeaf & 200 & 242 & 427 & 6 \\
Plane & 105 & 105 & 144 & 7 \\
Prox.Pha.Outln.AgeGrp. & 400 & 205 & 80 & 3 \\
Prox.PhalanxTW & 400 & 205 & 80 & 6 \\
SonyAIBORobotSurf.1 & 20 & 601 & 70 & 2 \\
SonyAIBORobotSurf.2 & 27 & 953 & 65 & 2 \\
SwedishLeaf & 500 & 625 & 128 & 15 \\
Symbols & 25 & 995 & 398 & 6 \\
ToeSegmentation1 & 40 & 228 & 277 & 2 \\
ToeSegmentation2 & 36 & 130 & 343 & 2 \\
TwoPatterns & 1000 & 4000 & 128 & 4 \\
TwoLeadECG & 23 & 1139 & 82 & 2 \\
Wafer & 1000 & 6164 & 152 & 2 \\
Wine & 57 & 54 & 234 & 2 \\
WordSynonyms & 267 & 638 & 270 & 25 \\
\bottomrule
\end{tabular}
    \label{tab:sampleData}
\end{table}

\subsubsection{Overview of Evaluation Methods}
We evaluate the classification performance using the following methods:
\begin{description}
    \item[\gls{ari}:] Similarity of learned and ground truth assignments \cite{10.1145/1553374.1553511}. For matched clustering partitions 
    \begin{equation*}
        \mathrm{RI} = (a+b)/\binom{N}{2}, \qquad \mathrm{ARI} =\frac{\mathrm{RI} - \mathbb{E} [\mathrm{RI}]}{\max(\text{RI})- \mathbb{E}[RI]]}
    \end{equation*}
    \item[\gls{nmi}:] The mutual information between the true class labels and the cluster assignments, normalized by the entropy of the true labels and the cluster assignments \cite{10.1145/1553374.1553511}.
        \begin{equation*}
        \text{NMI} = \frac{2 I(X; Y)}{H(X) + H(Y)}
    \end{equation*}
    where $X, Y$ are the true and assigned labels, $H(X)$ is the entropy of $X$, and $I(X; Y)$ is the mutual information between $X$ and $Y$.
    \item[Accuracy:] Proportion of correct predictions to a total number of predictions. 
    \item[F1:] Harmonic mean of the precision and recall. In the multiclass case, we take the macro average by calculating the metric for each label and computing their unweighted mean.
    \begin{equation*}
        F_1 = \frac{2 \times \mathrm{TP}}{2 \times \mathrm{TP} + \mathrm{FP} +\mathrm{FN}}
    \end{equation*}
    \item[Runtime:] Runtime of the algorithm in terms the user CPU time in the computational setting described in \ref{sec:exp}. If the method is GPU accelerated we report the user CPU/GPU time spent in the routine.
\end{description}

\subsubsection{Overview of Methods}
\label{apx:exp.ucr.baselines}
In addition to our method, we evaluate the performance of the following baseline methods
\begin{description}
    \item[$\ell_2$+KNN] K-Nearest Neighbors Classification with the distance metric as the Euclidean distance between two time-series treating the entire observation sequence as a high dimensional vector \cite{1053964}.
    \item[DTW+KNN] K-Nearest Neighbors Classification with the distance metric calculated according to \gls{dtw} \cite{Muller2007}, which aims to align the two given sequences by solving the following program
    \begin{equation*}
        \mathrm{DTW}_q(\mathbf{x}, \mathbf{x}^\prime) = \min_{\pi \in \mathcal{A}(\mathbf{x}, \mathbf{x}^\prime)} \langle A_\pi ,D_q(\mathbf{x}, \mathbf{x}^\prime)\rangle^{1/q}.
    \end{equation*}
    The set $\mathcal{A}(\mathbf{x}, \mathbf{x}^\prime)$ is the set of all admissible paths as represented by boolean matrices. Non-zero entries correspond to matching time series elements in the path. A path is admissible if the beginning and end of the time series are matched together, the sequence is monotone in both $i$ and $j$, and all entries appear at least once. We take $q=2$ as the Euclidean metric.
    \item[shapeDTW] Extension to \gls{dtw} scheme by incorporating point-wise local structures into the matching procedure \cite{ZHAO2018171}. Examples of such \textit{shape descriptors} include data itself, a rolling average of, a discrete wavelet transform, and a finite difference/derivative. Finally, the encoded sequences are then aligned by \gls{dtw} and used for nearest neighbor classification.
    \item[MR-Hydra] Combination of dictionary-based Multirocket and Hydra algorithms for time series classification, extracts and counts symbolic patterns using competing convolutional kernels \cite{Dempster:2023aa, MR}.
    \item[TS2Vec] Construct an encoder network for time series embedding based on hierarchical contrastive learning \cite{Yue_Wang_Duan_Yang_Huang_Tong_Xu_2022}. The discrimination is done both between sequences and within the sequences themselves. The encoder network consists of an input projection layer, a timestamp masking module, and a dilated convolutional module, and is optimized jointly with temporal and cross-sequence contrastive loss.
\end{description}

\subsubsection{Detailed Experimental Procedure}
We split our data into testing and training splits according to those given by the UCR repository. For the methods that directly perform classification (KNN, shapeDTW, Inception Time), we train on the test set and then report the the performance on the training set. In line with \cite{Yue_Wang_Duan_Yang_Huang_Tong_Xu_2022, NEURIPS2019_53c6de78}, to evaluate the classification performance on test set for methods which produce embeddings (TS2Vec and our method), we perform cross-validated grid search (based on $k=5$ folds, ) across \glspl{knn} with $k=\{2^i \mid i \in[0,4]\}$ neighbors or SVMs with RBF kernels with penalty values $c\in\{2^i \mid i \in [-10,15]\}\cup{\infty}$. For the KNN-based methods, we do the same grid search as outlined above. For our own method, we also grid search across parameters of $\lambda$ and report the performance for the best choice under rank constraint. To find the embedding, we run Algorithm \ref{alg:mp} for 256 iterations.

\subsection{Detailed results}
In Tables \ref{tab:resultbyDataset1} and \ref{tab:resultbyDataset2}, we present the classification performance for the discussed metrics for the evaluated methods for each of the tested UCR datasets.

\begin{table}

    \caption{Detailed results per UCR dataset (Part I)}
    
    \label{tab:resultbyDataset1}

\tiny
\centering
\begin{tabular}{lllllllllllll}
\toprule
Dataset                        & Method   & ARI    & NMI   & Acc.   & F1    & RT & Method        & ARI   & NMI   & Acc.   & F1    & RT     \\\cmidrule(lr){1-1} \cmidrule(lr){2-7} \cmidrule(lr){8-13}
ArrowHead                      & $\ell_2$+      & 0.482  & 0.453 & 0.800 & 0.800 & 0.007   &MR       & 0.629  & 0.585 & 0.863 & 0.863 & 3.183            \\
Beef                           & KNN    & 0.322  & 0.518 & 0.667 & 0.672 & 0.001   & Hydra    & 0.454  & 0.627 & 0.767 & 0.768 & 1.967            \\
BeetleFly                      &          & 0.219  & 0.344 & 0.750 & 0.733 & 0.001   &            & 0.621  & 0.619 & 0.900 & 0.899 & 1.532            \\
BirdChicken                    &          & -0.044 & 0.007 & 0.550 & 0.549 & 0.001   &            & 0.621  & 0.619 & 0.900 & 0.899 & 1.536            \\
Car                            &          & 0.403  & 0.477 & 0.733 & 0.737 & 0.003   &            & 0.830  & 0.860 & 0.933 & 0.933 & 4.023            \\
ChlorineConc.          &          & 0.231  & 0.157 & 0.650 & 0.610 & 0.633   &            & 0.472  & 0.373 & 0.789 & 0.753 & 57.36           \\
Coffee                         &          & 1.000  & 1.000 & 1.000 & 1.000 & 0.001   &            & 1.000  & 1.000 & 1.000 & 1.000 & 1.411            \\
DiatomsizeReduction            &          & 0.872  & 0.830 & 0.935 & 0.883 & 0.004   &            & 0.921  & 0.896 & 0.964 & 0.947 & 6.537            \\
Dist.Pha.Outln.AgeGrp.   &          & 0.190  & 0.224 & 0.626 & 0.613 & 0.019   &            & 0.383  & 0.404 & 0.770 & 0.775 & 4.794            \\
Dist.Pha.Outln.Correct    &          & 0.181  & 0.137 & 0.717 & 0.684 & 0.054   &            & 0.366  & 0.286 & 0.804 & 0.790 & 8.056            \\
ECG200                         &          & 0.571  & 0.445 & 0.880 & 0.868 & 0.004   &            & 0.667  & 0.542 & 0.910 & 0.902 & 1.765            \\
ECGFiveDays                    &          & 0.352  & 0.304 & 0.797 & 0.794 & 0.011   &            & 1.000  & 1.000 & 1.000 & 1.000 & 7.914            \\
GunPoint                       &          & 0.681  & 0.578 & 0.913 & 0.913 & 0.004   &            & 1.000  & 1.000 & 1.000 & 1.000 & 2.064            \\
Ham                            &          & 0.031  & 0.029 & 0.600 & 0.600 & 0.006   &            & 0.229  & 0.177 & 0.743 & 0.742 & 5.055            \\
Herring                        &          & -0.015 & 0.003 & 0.516 & 0.516 & 0.003   &            & 0.207  & 0.155 & 0.734 & 0.726 & 3.825            \\
Lightning2                     &          & 0.246  & 0.193 & 0.754 & 0.750 & 0.003   &            & 0.104  & 0.084 & 0.672 & 0.665 & 5.208            \\
Meat                           &          & 0.799  & 0.797 & 0.933 & 0.935 & 0.003   &            & 0.810  & 0.808 & 0.933 & 0.933 & 3.096            \\
Mid.Pha.Outln.AgeGrp.   &          & 0.055  & 0.026 & 0.519 & 0.443 & 0.021   &            & 0.092  & 0.071 & 0.591 & 0.491 & 4.599            \\
Mid.Pha.Outln.Correct    &          & 0.280  & 0.208 & 0.766 & 0.756 & 0.057   &            & 0.475  & 0.372 & 0.845 & 0.842 & 8.154            \\
Mid.PhalanxTW                &          & 0.379  & 0.367 & 0.513 & 0.382 & 0.020   &            & 0.383  & 0.433 & 0.513 & 0.339 & 5.382            \\
MoteStrain                     &          & 0.573  & 0.467 & 0.879 & 0.877 & 0.015   &            & 0.794  & 0.699 & 0.946 & 0.945 & 7.062            \\
OSULeaf                        &          & 0.298  & 0.383 & 0.521 & 0.525 & 0.023   &            & 0.921  & 0.919 & 0.963 & 0.956 & 10.34           \\
Plane                          &          & 0.919  & 0.943 & 0.962 & 0.963 & 0.005   &            & 1.000  & 1.000 & 1.000 & 1.000 & 2.505            \\
Prox.Pha.Outln.AgeGrp. &          & 0.492  & 0.422 & 0.785 & 0.693 & 0.027   &            & 0.662  & 0.564 & 0.868 & 0.797 & 5.489            \\
Prox.PhalanxTW              &          & 0.584  & 0.566 & 0.707 & 0.444 & 0.027   &            & 0.718  & 0.671 & 0.805 & 0.490 & 4.922            \\
SonyAIBORobotSurf.1          &          & 0.148  & 0.280 & 0.696 & 0.688 & 0.007   &            & 0.598  & 0.570 & 0.887 & 0.887 & 3.396            \\
SonyAIBORobotSurf.2          &          & 0.514  & 0.395 & 0.859 & 0.849 & 0.013   &            & 0.782  & 0.682 & 0.942 & 0.940 & 4.828            \\
SwedishLeaf                    &          & 0.629  & 0.761 & 0.789 & 0.782 & 0.109   &            & 0.950  & 0.965 & 0.976 & 0.977 & 10.78           \\
Symbols                        &          & 0.791  & 0.843 & 0.899 & 0.898 & 0.017   &            & 0.955  & 0.954 & 0.981 & 0.981 & 21.89           \\
ToeSegmentation1               &          & 0.126  & 0.095 & 0.680 & 0.675 & 0.006   &            & 0.832  & 0.782 & 0.956 & 0.956 & 4.760            \\
ToeSegmentation2               &          & 0.340  & 0.244 & 0.808 & 0.744 & 0.003   &            & 0.640  & 0.464 & 0.915 & 0.866 & 3.608            \\
TwoPatterns                    &          & 0.770  & 0.726 & 0.907 & 0.906 & 1.328   &            & 1.000  & 1.000 & 1.000 & 1.000 & 50.14           \\
TwoLeadECG                     &          & 0.244  & 0.217 & 0.747 & 0.741 & 0.015   &            & 0.993  & 0.983 & 0.998 & 0.998 & 6.182            \\
Wafer                          &          & 0.971  & 0.923 & 0.995 & 0.988 & 2.088   &            & 0.998  & 0.993 & 1.000 & 0.999 & 76.27           \\
Wine                           &          & 0.031  & 0.036 & 0.611 & 0.611 & 0.002   &            & 0.720  & 0.687 & 0.926 & 0.926 & 1.718            \\
WordSynonyms                   &          & 0.537  & 0.571 & 0.618 & 0.465 & 0.069   &            & 0.725  & 0.753 & 0.777 & 0.658 & 15.76           \\
\cmidrule(lr){1-1} \cmidrule(lr){2-7} \cmidrule(lr){8-13}
ArrowHead                      & DTW+   & 0.312  & 0.282 & 0.703 & 0.700 & 2.139   & TS2        & 0.480 & 0.462 & 0.794 & 0.794 & 81.08\\
Beef                           & KNN    & 0.276  & 0.490 & 0.633 & 0.629 & 1.158   & Vec        & 0.284 & 0.494 & 0.667 & 0.670 & 109.0\\
BeetleFly                      &          & 0.131  & 0.275 & 0.700 & 0.670 & 0.618   &               & 0.800 & 0.761 & 0.950 & 0.950 & 79.59\\
BirdChicken                    &          & 0.212  & 0.221 & 0.750 & 0.744 & 0.606   &               & 0.621 & 0.619 & 0.900 & 0.899 & 80.60\\
Car                            &          & 0.446  & 0.501 & 0.733 & 0.728 & 7.565   &               & 0.709 & 0.787 & 0.867 & 0.867 & 298.9\\
ChlorineConc.          &          & 0.231  & 0.154 & 0.648 & 0.607 & 247.313 &               & 0.432 & 0.333 & 0.764 & 0.730 & 2439\\
Coffee                         &          & 1.000  & 1.000 & 1.000 & 1.000 & 0.336   &               & 0.857 & 0.811 & 0.964 & 0.964 & 129.1\\
DiatomsizeReduction            &          & 0.938  & 0.921 & 0.967 & 0.942 & 3.272   &               & 0.968 & 0.952 & 0.984 & 0.973 & 87.44\\
Dist.Pha.Outln.AgeGrp.   &          & 0.389  & 0.368 & 0.770 & 0.763 & 1.804   &               & 0.272 & 0.277 & 0.705 & 0.699 & 2046\\
Dist.Pha.Outln.Correct    &          & 0.183  & 0.132 & 0.717 & 0.690 & 5.382   &               & 0.246 & 0.176 & 0.750 & 0.737 & 2882\\
ECG200                         &          & 0.280  & 0.192 & 0.770 & 0.749 & 0.468   &               & 0.540 & 0.417 & 0.870 & 0.858 & 463.5\\
ECGFiveDays                    &          & 0.286  & 0.252 & 0.768 & 0.763 & 1.848   &               & 0.991 & 0.979 & 0.998 & 0.998 & 80.84\\
GunPoint                       &          & 0.659  & 0.557 & 0.907 & 0.907 & 0.847   &               & 0.973 & 0.949 & 0.993 & 0.993 & 234.7\\
Ham                            &          & -0.005 & 0.003 & 0.467 & 0.467 & 12.116  &               & 0.210 & 0.168 & 0.733 & 0.733 & 526.1\\
Herring                        &          & -0.012 & 0.001 & 0.531 & 0.520 & 6.536   &               & 0.064 & 0.047 & 0.641 & 0.625 & 326.8\\
Lightning2                     &          & 0.537  & 0.480 & 0.869 & 0.864 & 8.898   &               & 0.318 & 0.252 & 0.787 & 0.783 & 301.2\\
Meat                           &          & 0.799  & 0.797 & 0.933 & 0.935 & 4.497   &               & 0.687 & 0.714 & 0.883 & 0.883 & 300.1\\
Mid.Pha.Outln.AgeGrp.   &          & 0.024  & 0.022 & 0.500 & 0.411 & 2.066   &               & 0.038 & 0.029 & 0.519 & 0.426 & 2042\\
Mid.Pha.Outln.Correct    &          & 0.153  & 0.109 & 0.698 & 0.691 & 5.663   &               & 0.385 & 0.297 & 0.811 & 0.808 & 3055\\
Mid.PhalanxTW                &          & 0.380  & 0.368 & 0.506 & 0.374 & 1.996   &               & 0.420 & 0.404 & 0.545 & 0.396 & 1997\\
MoteStrain                     &          & 0.448  & 0.351 & 0.835 & 0.834 & 0.900   &               & 0.528 & 0.424 & 0.863 & 0.862 & 86.43\\
OSULeaf                        &          & 0.309  & 0.392 & 0.591 & 0.588 & 50.818  &               & 0.644 & 0.671 & 0.822 & 0.799 & 1071\\
Plane                          &          & 1.000  & 1.000 & 1.000 & 1.000 & 1.172   &               & 1.000 & 1.000 & 1.000 & 1.000 & 541.7\\
Prox.Pha.Outln.AgeGrp. &          & 0.504  & 0.430 & 0.805 & 0.716 & 2.696   &               & 0.506 & 0.437 & 0.780 & 0.689 & 2051\\
Prox.PhalanxTW              &          & 0.644  & 0.587 & 0.756 & 0.511 & 2.673   &               & 0.674 & 0.628 & 0.771 & 0.562 & 2052\\
SonyAIBORobotSurf.1          &          & 0.200  & 0.316 & 0.725 & 0.721 & 0.309   &               & 0.588 & 0.571 & 0.884 & 0.883 & 84.73\\
SonyAIBORobotSurf.2          &          & 0.435  & 0.324 & 0.831 & 0.817 & 0.571   &               & 0.671 & 0.584 & 0.910 & 0.907 & 125.5\\
SwedishLeaf                    &          & 0.639  & 0.770 & 0.792 & 0.787 & 25.957  &               & 0.875 & 0.916 & 0.936 & 0.937 & 2573\\
Symbols                        &          & 0.891  & 0.913 & 0.950 & 0.949 & 21.459  &               & 0.928 & 0.936 & 0.969 & 0.969 & 132.7\\
ToeSegmentation1               &          & 0.293  & 0.260 & 0.772 & 0.762 & 3.704   &               & 0.816 & 0.723 & 0.952 & 0.952 & 213.5\\
ToeSegmentation2               &          & 0.398  & 0.249 & 0.838 & 0.764 & 2.904   &               & 0.714 & 0.631 & 0.931 & 0.899 & 170.8\\
TwoPatterns                    &          & 1.000  & 1.000 & 1.000 & 1.000 & 330.318 &               & 0.973 & 0.964 & 0.990 & 0.990 & 5216\\
TwoLeadECG                     &          & 0.654  & 0.564 & 0.904 & 0.904 & 0.901   &               & 0.958 & 0.918 & 0.989 & 0.989 & 87.20\\
Wafer                          &          & 0.867  & 0.748 & 0.980 & 0.944 & 720.459 &               & 0.942 & 0.868 & 0.991 & 0.976 & 5386\\
Wine                           &          & 0.003  & 0.016 & 0.574 & 0.574 & 0.860   &               & 0.339 & 0.271 & 0.796 & 0.796 & 303.5\\
WordSynonyms                   &          & 0.575  & 0.600 & 0.649 & 0.533 & 66.754  &               & 0.349 & 0.421 & 0.522 & 0.309 & 1407\\
\bottomrule
\end{tabular}
\end{table}

\begin{table}    

\caption{Detailed results per UCR dataset (Part II)}

\label{tab:resultbyDataset2}
\tiny
    \centering
\begin{tabular}{lllllllllllll}
\toprule
Dataset                        & Method   & ARI    & NMI   & Acc.   & F1    & RT & Method        & ARI   & NMI   & Acc.   & F1    & RT     \\\cmidrule(lr){1-1} \cmidrule(lr){2-7} \cmidrule(lr){8-13}
ArrowHead                      & shape    & 0.521  & 0.492 & 0.817 & 0.818 & 0.672   & Ours&0.336 & 0.306 & 0.720 & 0.720 & 136.7         \\
Beef                           & DTW      & 0.322  & 0.518 & 0.667 & 0.672 & 0.214   &     &0.453 & 0.654 & 0.733 & 0.736 & 70.59         \\
BeetleFly                      &          & 0.219  & 0.344 & 0.750 & 0.733 & 0.107   &     &1.000 & 1.000 & 1.000 & 1.000 & 51.50         \\
BirdChicken                    &          & -0.044 & 0.007 & 0.550 & 0.549 & 0.106   &     &1.000 & 1.000 & 1.000 & 1.000 & 51.92         \\
Car                            &          & 0.560  & 0.585 & 0.817 & 0.815 & 1.077   &     &0.372 & 0.457 & 0.667 & 0.648 & 173.1         \\
ChlorineConc.                  &          & 0.199  & 0.133 & 0.628 & 0.587 & 120.800 &     &0.537 & 0.414 & 0.811 & 0.782 & 2252          \\
Coffee                         &          & 1.000  & 1.000 & 1.000 & 1.000 & 0.105   &     &1.000 & 1.000 & 1.000 & 1.000 & 39.92         \\
DiatomsizeReduction            &          & 0.921  & 0.890 & 0.958 & 0.921 & 0.849   &     &0.818 & 0.865 & 0.882 & 0.704 & 331.7         \\
Dist.Pha.Outln.AgeGrp.         &          & 0.209  & 0.251 & 0.633 & 0.615 & 1.584   &     &0.323 & 0.402 & 0.741 & 0.748 & 103.9         \\
Dist.Pha.Outln.Correct         &          & 0.188  & 0.140 & 0.721 & 0.690 & 4.733   &     &0.316 & 0.233 & 0.783 & 0.772 & 158.0         \\
ECG200                         &          & 0.541  & 0.420 & 0.870 & 0.860 & 0.397   &     &0.668 & 0.554 & 0.910 & 0.904 & 43.80         \\
ECGFiveDays                    &          & 0.705  & 0.605 & 0.920 & 0.920 & 1.279   &     &0.765 & 0.666 & 0.937 & 0.937 & 306.1         \\
GunPoint                       &          & 0.845  & 0.761 & 0.960 & 0.960 & 0.517   &     &0.845 & 0.761 & 0.960 & 0.960 & 70.03         \\
Ham                            &          & 0.031  & 0.029 & 0.600 & 0.600 & 2.714   &     &0.160 & 0.124 & 0.705 & 0.704 & 265.6         \\
Herring                        &          & -0.012 & 0.006 & 0.531 & 0.531 & 1.129   &     &0.176 & 0.128 & 0.719 & 0.688 & 162.0         \\
Lightning2                     &          & 0.358  & 0.299 & 0.803 & 0.797 & 1.345   &     &0.399 & 0.320 & 0.820 & 0.817 & 230.8         \\
Meat                           &          & 0.799  & 0.797 & 0.933 & 0.935 & 0.860   &     &0.856 & 0.841 & 0.950 & 0.950 & 133.3         \\
Mid.Pha.Outln.AgeGrp.          &          & 0.053  & 0.022 & 0.513 & 0.432 & 1.722   &     &0.184 & 0.143 & 0.649 & 0.523 & 106.6         \\
Mid.Pha.Outln.Correct          &          & 0.281  & 0.207 & 0.766 & 0.759 & 5.110   &     &0.410 & 0.324 & 0.821 & 0.813 & 177.9         \\
Mid.PhalanxTW                  &          & 0.357  & 0.360 & 0.487 & 0.361 & 1.740   &     &0.362 & 0.436 & 0.591 & 0.334 & 107.5         \\
MoteStrain                     &          & 0.573  & 0.467 & 0.879 & 0.877 & 0.804   &     &0.212 & 0.175 & 0.731 & 0.731 & 243.3         \\
OSULeaf                        &          & 0.316  & 0.411 & 0.566 & 0.567 & 10.754  &     &0.600 & 0.625 & 0.810 & 0.798 & 584.7         \\
Plane                          &          & 0.937  & 0.961 & 0.971 & 0.972 & 0.644   &     &1.000 & 1.000 & 1.000 & 1.000 & 74.50         \\
Prox.Pha.Outln.AgeGrp.         &          & 0.482  & 0.399 & 0.780 & 0.688 & 2.244   &     &0.681 & 0.584 & 0.883 & 0.808 & 119.3         \\
Prox.PhalanxTW                 &          & 0.585  & 0.565 & 0.702 & 0.426 & 2.240   &     &0.752 & 0.728 & 0.834 & 0.436 & 118.7         \\
SonyAIBORobotSurf.1            &          & 0.206  & 0.333 & 0.729 & 0.724 & 0.282   &     &0.619 & 0.572 & 0.894 & 0.893 & 102.1         \\
SonyAIBORobotSurf.2            &          & 0.589  & 0.468 & 0.885 & 0.876 & 0.560   &     &0.767 & 0.655 & 0.938 & 0.934 & 139.3         \\
SwedishLeaf                    &          & 0.697  & 0.806 & 0.830 & 0.827 & 16.027  &     &0.805 & 0.864 & 0.899 & 0.897 & 441.6         \\
Symbols                        &          & 0.823  & 0.864 & 0.918 & 0.917 & 4.945   &     &0.903 & 0.917 & 0.956 & 0.956 & 1127          \\
ToeSegmentation1               &          & 0.221  & 0.181 & 0.737 & 0.728 & 1.136   &     &0.678 & 0.580 & 0.912 & 0.912 & 203.4         \\
ToeSegmentation2               &          & 0.486  & 0.379 & 0.862 & 0.809 & 0.780   &     &0.655 & 0.471 & 0.923 & 0.868 & 158.2         \\
TwoPatterns                    &          & 0.908  & 0.870 & 0.965 & 0.964 & 204.000 &     &0.679 & 0.145 & 0.143 & 0.514 & 1719          \\
TwoLeadECG                     &          & 0.484  & 0.438 & 0.848 & 0.846 & 0.742   &     &0.993 & 0.981 & 0.998 & 0.998 & 240.5         \\
Wafer                          &          & 0.977  & 0.936 & 0.996 & 0.991 & 373.891 &     &0.939 & 0.858 & 0.991 & 0.975 & 3329          \\
Wine                           &          & 0.016  & 0.025 & 0.593 & 0.593 & 0.300   &     &0.150 & 0.128 & 0.704 & 0.702 & 62.29         \\
WordSynonyms                   &          & 0.578  & 0.600 & 0.639 & 0.487 & 20.963  &     &0.246 & 0.325 & 0.395 & 0.220 & 764.7         \\
\bottomrule
\end{tabular}

\end{table}

\subsection{Natural language embedding}
\label{apx:exp.lang}
To acquire the data, we retrieved the raw text of \textit{Alice's Adventures in Wonderland} and \textit{Through the Looking Glass} from Project Gutenberg \footnote{\url{https://www.gutenberg.org}}. For the paper abstracts, we used the training portion of the \texttt{ML-ArXiv-Papers} dataset \footnote{\url{https://www.kaggle.com/datasets/Cornell-University/arxiv}}. For each dataset, we stripped all non ASCII characters and uncommon punctuation 
(\texttt{<},\texttt{>}, \texttt{`}, \texttt{=}, \texttt{|}, \texttt{?}, \texttt{\&}, \texttt{[}, \texttt{]}, \texttt{*}, \texttt{~}, \texttt{!}, \texttt{\#}, \texttt{@}, and \texttt{"}).

After acquiring the data, we then encoded using a Huffman tree with $n=4$ symbols derived from the frequency of letters in our corpora. We treated each abstract as a document and considered 500-character chunks of the two books. We rejected abstracts containing less than 500 words. After encoding the sequences using the Huffman code, we cut off each sequence at 1000 coded symbols and rejected all sequences less than this length after coding. This left us with $n=228$ samples from  ``Alice's Adventures in Wonderland'', $n=316$ samples from ``Through the Looking Glass'', and $n=600$ machine learning-related ArXiv abstracts.

To learn the embedding, we use the method described in Appendix \ref{apx:nbfom} and grid searched across values of $\lambda$ for $512$ steps using the softmax link function described in Section \ref{sec:SEQ}.

\subsection{Gene sequence embedding}
\label{apx:gene}
\paragraph{Data acquisition and processing}
In line with \cite{10.1371/journal.pone.0261531}, we downloaded viral genome sequences for two different kinds of human viruses: \textit{Influenza A virus} and  \textit{Dengue virus}. We consider different strains of each virus in addition to the species as a whole. We provide a textual description below. In Table \ref{tab:viralstat}, we provide summary statistics, including the number of sequences in the strain, the average and standard deviation of the sequence lengths, and the length of the shortest and longest sequences in the strains.

\begin{description}
    \item[Influenza A] The \textit{Influenza A virus} genome data ($n=949$) is acquired from the NCBI Influenza Virus Resource \cite{Bao2008-oq}. We consider the genome of segment 6, which encodes the neuraminidase protein, and include sequence samples belonging to subtypes H1N1, H2N2, H5N1, H7N3, and H7N9. 
    \item[Denuge] We consider $n=1562$ full \textit{Dengue virus} genomes downloaded from the NCBI Virus Variation Resource \cite{Hatcher2017-hh}. We consider all four subtypes of the virus DENV-1, DENV-2, DENV-3, and DENV-4. 
\end{description}

We encoded all the sequences as four-channel signals via one-hot encoding, with each nucleotide (A,C,T,G) corresponding to one of the channels. In the case we encounter incompletely specified bases in the nucleic acid sequences \cite{10.1042/bj2290281}, we give equiprobable weights to the possible corresponding nucleotides.

\paragraph{Learning procedure}
Same as the natural language case, we represent the data as a four channel signal and adopt the softmax activation scheme as described in Section \ref{sec:SEQ}. Since the sequences are of considerable and variable length ($>1000$ nucleotides, see Table \ref{tab:viralstat}), we adopt a stochastic estimation to  \eqref{eqn:MonotoneVI} by randomly sampling length $G=800$ sub-windows from each of sequences. We take the sample average for each of the sub-window observations similar to \eqref{eqn:vi.ssa}. We run Algorithm \ref{alg:mp} for $N=1024$ iterations, using the stochastic approximation described above and grid searching across values of $\lambda$. To produce Figures \ref{fig:subfigb} and \ref{fig:subfigc}, we took the learned representations and projected them into two dimensions via UMAP.

\begin{table}
    \centering
    \caption{Statistics for selected viral genomes.}
    \label{tab:viralstat}
    \begin{tabular}{lllllll}
    \toprule
    Virus & Strain & Count & \multicolumn{4}{c}{Sequence Length} \\\cmidrule(lr){4-7}
    &&& Avg. & Std. & Min. & Max.\\
    \midrule
    Influenza-A & H5N1 & 188 & 1368.521 & (21.682) & 1350 & 1457\\
                & H1N1 & 191 & 1421.0 & (15.25) & 1350 & 1468\\
                & H7N9 & 190 & 1403.521 & (12.048) & 1389 & 1444\\
                & H2N2 & 187 & 1430.053 & (17.87) & 1376 & 1467\\
                & H7N3 & 193 & 1423.15 & (21.537) & 1345 & 1468\\
                \cmidrule(lr){2-7}
                &\textbf{Total} & 949 & 1409.326 & (28.512) & 1345 & 1468\\
    \midrule
    Dengue & DENV-1 & 409 & 10577.812 & (194.4) & 10176 & 10821\\
           & DENV-2 & 409 & 10592.504 & (196.308) & 10173 & 10991\\
           & DENV-3 & 408 & 10614.137 & (132.911) & 10173 & 10810\\
           & DENV-4 & 407 & 10452.469 & (205.208) & 10161 & 10772\\
           \cmidrule(lr){2-7}
                &\textbf{Total} & 1633 & 10559.328 & (194.74) & 10161 & 10991\\
    \bottomrule
    \end{tabular}
\end{table}

\end{document}